%% file: main.tex
\documentclass{article}

\input{preamble}

\title{Transect: Retaining Observability for Long-Horizon LLM Agent Evaluations}

\renewcommand{\shorttitle}{Transect}

\author{
\textbf{Toby D. Pilditch}\textsuperscript{1}\thanks{Corresponding author: \href{mailto:toby.pilditch@dsit.gov.uk}{toby.pilditch@dsit.gov.uk}.}
\qquad \textbf{Konstantinos Voudouris}\textsuperscript{1}\\[0.4em]
\textbf{Alexandra Abbas}\textsuperscript{2}\qquad \textbf{Cozmin Ududec}\textsuperscript{1}\\[0.9em]
{\normalfont\textsuperscript{1}UK AI Security Institute}\\[0.3em]
{\normalfont\textsuperscript{2}Meridian Labs}
}

\date{}

\begin{document}

\maketitle

\begin{abstract}
\input{sections/00_abstract}
\end{abstract}

\keywords{LLM evaluation \and transcript analysis \and observability \and
  AI safety \and agentic evaluation \and expert elicitation}

\input{sections/01_introduction}
\input{sections/02_Walkthrough_intro}
\input{sections/02b_Walkthrough_case}
\input{sections/03_conclusion}

\bibliographystyle{unsrtnat}   
\bibliography{references}

\appendix
\input{sections/04_appendix}

\end{document}

%% file: preamble.tex
\usepackage{arxiv}              

\usepackage[utf8]{inputenc}
\usepackage[T1]{fontenc}
\usepackage{microtype}          
\usepackage{nicefrac}           

\usepackage{amsmath}
\usepackage{amssymb}
\usepackage{amsfonts}

\usepackage{graphicx}
\graphicspath{{figures/}}
\usepackage{booktabs}           
\usepackage{tabularx}
\usepackage{array}
\usepackage{multirow}
\usepackage{float}
\usepackage{subcaption}

\usepackage{enumitem}
\usepackage{xcolor}

\usepackage{tikz}
\usetikzlibrary{arrows}
\usetikzlibrary{arrows.meta, positioning, fit}
\definecolor{aisinavy}{HTML}{0E2841}
\definecolor{aisisteel}{HTML}{104D6A}
\definecolor{aisiteal}{HTML}{09A4A7}
\definecolor{aisislate}{HTML}{8997A6}
\definecolor{aisired}{HTML}{F90D29}
\definecolor{aisimute}{HTML}{5F6B7A}    
\definecolor{aisicrimson}{HTML}{C50016} 
\definecolor{aisipanel}{HTML}{EEF1F4}   
\definecolor{aisiwarm}{HTML}{FDF0F1}    
\usepackage[numbers,sort&compress]{natbib}

\usepackage[colorlinks=true,
            linkcolor=blue,
            citecolor=blue!55!black,
            urlcolor=blue!55!black,
            filecolor=blue!55!black]{hyperref}
\usepackage{url}
\usepackage[capitalise,noabbrev]{cleveref}

\newif\ifdraftmode
\draftmodetrue                  

\newcommand{\transect}{Transect}

\newcommand{\evalfile}{\texttt{.eval}}

%% file: sections/00_abstract.tex
Frontier AI evaluations increasingly use open-ended, agentic, long-horizon tasks whose transcripts can span hundreds of pages of outputs and actions from complex multi-agent networks. The \textit{observability envelope}---the range of what evaluators can reliably infer about an agent's behaviours---is therefore narrowing. Language model assistants can help classify and interpret agent behaviour but also afford human evaluators significant analytical degrees of freedom, threatening the reproducibility and auditability of language-model-based transcript analysis. \transect{} is an open source package built on Inspect Scout to help evaluators understand how a long agent run unfolded, identify behaviour worth investigating, and check interpretations against the transcript. Users specify task context and behavioural vocabulary in a reusable evaluation-family configuration, with judge models and analysis settings supplied separately. \transect{}'s navigable reports align recorded events, token use, sub-agent activity, and model-generated behavioural labels on a common turn-based timeline. Reviewers can quickly grasp a run's narrative, trace any label or event to its source turns, and export the underlying data tables for cross-run analysis. We demonstrate the workflow on an AI R\&D evaluation that generated almost 13 million tokens, dividing the agents' work into behavioural phases aligned with research-skill classifications, sub-agent delegations and interactions, and token use. The combined view shows a focus on operational work and manuscript production, with little evidence of a sustained hypothesis generation stage---arguably a necessary component for high-quality scientific outputs. \transect{}'s flexible, customisable transcript-analysis pipeline will enable evaluators to keep pace with longer, more complex, more frequent AI evaluations while supporting scientific rigour, transparency, and reproducibility.

%% file: sections/01_introduction.tex
\section{Introduction}
\label{sec:intro}

Three shifts are reshaping frontier AI evaluation. First, static benchmarks continue to saturate, limiting their usefulness for distinguishing the capabilities of frontier AI systems \citep{zhou2026general}.\footnote{Note that henceforth we use the term \textit{AI model} to refer to the static set of network weights, \textit{AI agent} to refer to a model that has access to a set of tools in an environment and can perform actions with them, and \textit{AI system} to refer to the general infrastructure around and including that agent, including synchronous monitors, action classifiers, and multi-agent harnesses.} Evaluation has therefore moved towards open-ended, agentic, long-horizon tasks in which AI agents plan, use tools, delegate to sub-agents, and act over many steps towards a goal, producing millions, sometimes \textit{billions}, of tokens along the way \citep{kapoor2026openworld,yehudai2025survey,mohammadi2025survey}. Second, increasingly capable agents are arriving more frequently, leaving less time to scrutinise each system before the next appears, and agent capabilities can improve substantially between model releases \citep{folkerts2026cyber}. Third, an agent's measured performance can depend substantially on its token budget and opportunities to revise its work \citep{mcfadyen2026inferencecomputeshapesfrontier,aisi2026compute}. Evaluations are consequently becoming longer, more frequent, and more compute-intensive, imposing a growing burden on the human evaluators monitoring progress at the frontier.

We call the range of attributes an evaluator can reliably see or infer from an evaluation its \textbf{observability envelope}. This includes what an agent can do under the evaluation conditions, how dependably it performs, where and how it fails, and what propensities or dispositions it displays. The envelope depends on the evaluation apparatus as a whole: the transcript, the details of the AI system being evaluated (e.g., harness design), the transcript analysis methods, and those interpreting the evaluation evidence. There are at least four pressures on this observability envelope, increasing the risk that evaluators draw unjustified conclusions from transcripts. 

First, \textbf{within-run opacity}: a final score summarises performance on a long-horizon task but obscures discarded strategies, course corrections, recoveries, unexpected behaviors, and attempts to game the evaluation \citep{aisi2026cheating}. Second, a \textbf{human attention bottleneck}: reviewing hundreds or thousands of pages of agent activity requires scarce subject-matter expertise and the ability to connect and co-interpret complex events at distant points in a transcript \citep{ukaisi2026securityincident}. Reviewer time cannot grow in proportion to transcript volume and model release cadence \citep{reizinger2026skills,liu2026lasthuman}. Third, \textbf{insufficient statistical power}: long agent evaluations are costly in time and resources, so evaluators can afford relatively few repetitions per agent and setup, leaving measurements based on final scores alone vulnerable to task variance, stochasticity, and isolated qualitative anecdotes \citep{luettgau2025hibayes,nist2026toolbox}. Fourth, \textbf{setup attribution uncertainty}: observed performance is determined by both the model and the system it is embedded in \citep{zhang2026stop,yao2026harnessbench}. Without observing an agent's entire behavioural trajectory, it can be hard to assess the construct validity of an evaluation---whether the evaluation measures the intended capability rather than an artefact of its instruments or task setup \citep{cronbach1955construct}. 

Multi-agent evaluations further intensify these pressures on observability. Running agents in parallel can increase transcript volume within the same time budget, leaving more activity for reviewers to inspect. Interactions across agents and shared resources can make the course of the evaluation harder to follow. If an evaluation uses more total compute, fewer repetitions may be affordable within a fixed compute budget. Additional choices about multi-agent delegation and coordination can also make it harder to separate the effects of the model and its evaluation setup.

The evaluation community is aware of these pressures and has been building tooling to maintain and widen the observability envelope. Inspect AI and Inspect Scout provide primitives for recording, scanning, and querying evaluation traces, including with large language models \citep{inspectai,inspectscout}, while platforms such as Docent support interactive exploration of behavioural patterns across agent transcripts \citep{transluce_docent}. Another prominent approach is to use language model assistants with custom judging rubrics to aid transcript analysis \citep{metr2026incidentinvestigation}. However, such approaches introduce substantial researcher degrees of freedom, such as the judge model, rubric, unit of analysis (tool calls, chain of thought, chains of actions), and statistical analyses performed on top of these judgments. Analysts must therefore navigate a garden of forking paths in which defensible-seeming decisions accumulate into potentially unreproducible findings \citep{gelman2016statistical,steegen2016increasing,silberzahn2018many}. Model-assisted analysis can also make the evaluation \emph{more} opaque rather than less: a judge produces yet more tokens to interpret, can overstate confidence, and its errors can be difficult to verify \citep{bavaresco2025llms,norman2026reliability}. If findings about agent behaviour are to be trusted, these methods must be reproducible and the effect of arbitrary analytic choices must be measured. Otherwise, the tools intended to widen the observability envelope risk narrowing it further, replacing unread transcripts with unverifiable claims.

We present \textbf{\transect{}}, a transcript analysis package designed to address within-run opacity and the human attention bottleneck. Its shared, source-linked account helps evaluators understand how a run unfolded, identify behaviour worth investigating, and check their interpretations against the transcript. Reviewers can also use this account to investigate setup attribution uncertainty, although separating the effects of the model and its evaluation setup requires further evidence. Its structured outputs also support comparisons across runs, though they do not resolve insufficient statistical power when too few independent evaluations are available.

The analyst supplies a reusable configuration for the evaluation family detailing the information they need to extract, along with LLM judge rubrics and model specifications that will help them do so. Built on Inspect Scout \citep{inspectscout}, \transect{} analyses Inspect evaluation logs \cite{inspectai} or supported OpenClaw exports \cite{openclaw}. It extracts recorded events and uses the selected judges to classify activities, producing a report organised along a shared timeline and structured tables for further analysis. Reviewers can follow the course of the evaluation, examine activities alongside token use and sub-agent deployments, and follow labels and events directly back to the source in the transcript without assembling separate scanner outputs themselves. This connects the overall account of an evaluation to the details needed to investigate an unexpected behaviour, a possible safety incident, or a constraint imposed by the evaluation setup.

Reproducible analysis requires retaining the source transcripts, run selection, family configuration, judging settings, custom analysis code and stored results, and recording software and model versions where available (\Cref{app:reproducibility}). Independent evaluators can re-examine stored results without new judge calls, although rerunning the judges may produce different labels. The shared judging machinery records judge identities and decisions, with diagnostics of within-model stability or between-model agreement where the judging regime supports them. Of course, this information alone does not establish whether the labels are correct. These records and checks support \transect{}'s aim of scaling oversight responsibly: increasing analytical capacity while preserving the means to assess its findings.

The remainder of the paper describes \transect{}'s pipeline and design principles (\Cref{sec:transect}), then illustrates the workflow on a long-horizon AI R\&D shadow evaluation with extensive sub-agent activity (\Cref{sec:cases}). The case examines research activities and token expenditure, uses custom scanners to trace shared-file accesses across agents, and combines scanner outputs into additional report views. A full technical reference appears in \Cref{app:package}, and the implementation is openly available at \url{https://github.com/AI-Safety-Institute/transect}.

%% file: sections/02_Walkthrough_intro.tex
\section{Transect}
\label{sec:transect}


\transect{} turns an evaluation transcript into a navigable account of the agent's activities with direct links to their source. We describe the pipeline (\Cref{sec:transect:pipeline}) and the principles guiding its configuration and interpretation (\Cref{sec:transect:principles}), then work through one AI R\&D run to show how a reviewer can use the report (\Cref{sec:cases}).

\subsection{The pipeline}
\label{sec:transect:pipeline}
\transect{} builds on Inspect Scout's transcript and scanner primitives \citep{inspectscout} (\Cref{app:package}). The pipeline starts with a completed run transcript and a \textbf{Spec}: a reusable configuration for an evaluation \textit{family} (e.g., a cyber-security range or an AI R\&D evaluation). A Spec is used for every evaluation family, whether or not the analysis calls language models. It can contain task context, activity definitions, sub-agent role categories where relevant, and user-defined content for custom analyses. Its contents vary with the task and the user's questions: sub-agent categories, for example, are needed only when classifying sub-agent roles. The package includes skills for coding assistants to guide Spec authoring and refinement. The user starts from the evaluation brief and setup materials, identifies missing information, and seeks clarification from evaluation designers or subject-matter experts. For classification, the user supplies the categories and definitions, and language model scanners use them to label the transcript. The user can also add custom structural scanners to extract recorded information and custom judged scanners to classify behaviour (\Cref{app:enrichment}).

The broader pipeline configuration also includes separately supplied analysis settings, such as judge models and whether and how they may interact (see \Cref{app:pipeline}). \transect{} supports one pass by a single model (solo), repeated passes by that model ($k$-rolls), or a cohort of models making separate judgments (\Cref{app:trust}). These choices specify how to analyse the run, while the Spec describes evaluation content and context. \transect{} retains the supplied judging settings in its stored scanner configuration, alongside records of the judging regime and results, for later inspection (\Cref{app:reproducibility}). \transect{} returns a \texttt{TransectResults} object containing \textbf{dataframes}, the structured tables underlying a navigable report (\Cref{app:datamodel}).

The report helps reviewers locate episodes that warrant inspection, compare token use across activities, and investigate how recorded events relate to changes in behaviour. For example, reviewers can examine whether a shift in activity occurs near a context compaction and follow both back to the transcript. These features address the \textbf{human attention bottleneck} and \textbf{within-run opacity}, and help investigate how the evaluation setup shaped a run (\Cref{sec:intro}).

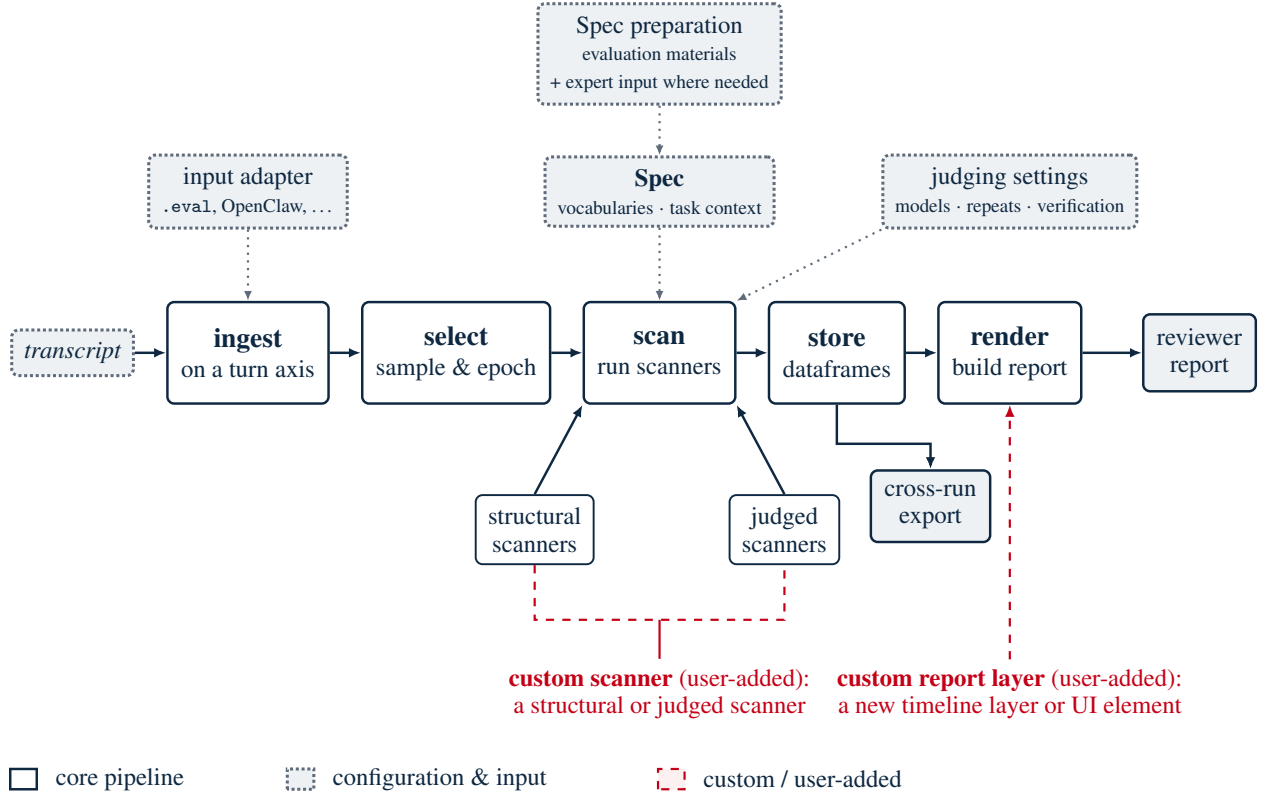
\begin{figure}[h]
\centering

\resizebox{\textwidth}{!}{%
\begin{tikzpicture}[
  font=\footnotesize,
  core/.style={draw=aisinavy, line width=0.9pt, rounded corners=2.5pt, fill=white,
               text=aisinavy, align=center, inner sep=5pt, minimum height=1.30cm},
  deliv/.style={draw=aisinavy, line width=0.9pt, rounded corners=2.5pt, fill=aisipanel,
                text=aisinavy, align=center, inner sep=5pt},
  comp/.style={draw=aisinavy, line width=0.7pt, rounded corners=2.5pt, fill=white,
               text=aisinavy, align=center, inner sep=4.5pt},
  cfg/.style={draw=aisimute, densely dotted, line width=1.1pt, rounded corners=2.5pt,
              fill=aisipanel, text=aisinavy, align=center, inner sep=4.5pt},
  input/.style={draw=aisimute, densely dotted, line width=1.1pt, rounded corners=2.5pt,
                fill=aisipanel, text=aisinavy, align=center, inner sep=4.5pt,
                font=\footnotesize\itshape},
  arr/.style={-{Latex[length=4.4pt,width=3.6pt]}, aisinavy, line width=0.9pt},
  clead/.style={-{Latex[length=3.6pt,width=3pt]}, aisimute, line width=0.8pt, dotted},
  carr/.style={aisicrimson, line width=0.8pt, dashed},
  carrow/.style={-{Latex[length=4.4pt,width=3.6pt]}, aisicrimson, line width=0.8pt, dashed},
]
\node[input] (tx)                       {transcript};
\node[core]  (ing) [right=4mm of tx]  {{\normalsize\bfseries ingest}\\[1pt]on a turn axis};
\node[core]  (sel) [right=4mm of ing] {{\normalsize\bfseries select}\\[1pt]sample \& epoch};
\node[core]  (scn) [right=4mm of sel] {{\normalsize\bfseries scan}\\[1pt]run scanners};
\node[core]  (df)  [right=4mm of scn] {{\normalsize\bfseries store}\\[1pt]dataframes};
\node[core]  (out) [right=4mm of df]  {{\normalsize\bfseries render}\\[1pt]build report};
\draw[arr] (tx) -- (ing);  \draw[arr] (ing) -- (sel);  \draw[arr] (sel) -- (scn);
\draw[arr] (scn) -- (df);  \draw[arr] (df) -- (out);
\node[deliv] (rep) at ([xshift=15mm]out.east) {reviewer\\report};
\node[deliv] (exp) at ([xshift=12mm, yshift=-13mm]df.south) {cross-run\\export};
\draw[arr] (out.east) -- (rep.west);
\draw[arr] (df.south) -- ++(0,-5mm) -| (exp.north);
\node[cfg] (spec)  at ([yshift=14mm]scn.north)  {\textbf{Spec}\\{\scriptsize vocabularies $\cdot$ task context}};
\node[cfg] (judge) at (out |- spec) {judging settings\\{\scriptsize models $\cdot$ repeats $\cdot$ verification}};
\node[cfg] (cal)   at ([yshift=13mm]spec.north) {Spec preparation\\{\scriptsize evaluation materials}\\{\scriptsize + expert input where needed}};
\node[cfg] (adapt) at (ing |- spec) {input adapter\\{\scriptsize \texttt{.eval}, OpenClaw, \dots}};
\draw[clead] (adapt.south) -- (ing.north);
\draw[clead] (spec.south)  -- (scn.north);
\draw[clead] (judge.south west) -- (scn.north east);
\draw[clead] (cal.south)   -- (spec.north);
\node[comp] (struct) at ([xshift=-16mm, yshift=-16mm]scn.south) {structural\\scanners};
\node[comp] (judged) at ([xshift= 16mm, yshift=-16mm]scn.south) {judged\\scanners};
\draw[arr] (struct.north) -- (scn.south west);
\draw[arr] (judged.north) -- (scn.south east);
\coordinate (bl) at ([yshift=-7mm]struct.south);
\coordinate (br) at ([yshift=-7mm]judged.south);
\draw[carr] (struct.south) -- (bl) -- (br) -- (judged.south);
\draw[aisicrimson, line width=0.8pt] (scn |- bl) -- ([yshift=-5mm]scn |- bl);
\node[text=aisicrimson, align=center] (lbl) at ([yshift=-9.5mm]scn |- bl)
  {\textbf{custom scanner} (user-added):\\a structural or judged scanner};
\node[text=aisicrimson, align=center] (rlbl) at (out |- lbl)
  {\textbf{custom report layer} (user-added):\\a new timeline layer or UI element};
\draw[carrow] (rlbl.north) -- (out.south);

\coordinate (lg) at ([yshift=-6.5mm]lbl.south);
\begin{scope}[shift={(tx.west |- lg)}]
  \fill[white]    (0, -0.13) rectangle (0.34, 0.13);
  \draw[aisinavy, line width=0.9pt] (0, -0.13) rectangle (0.34, 0.13);
  \node[anchor=west, text=aisinavy] at (0.46, 0) {core pipeline};
  \fill[aisipanel] (3.55, -0.13) rectangle (3.89, 0.13);
  \draw[aisimute, densely dotted, line width=1.1pt] (3.55, -0.13) rectangle (3.89, 0.13);
  \node[anchor=west, text=aisinavy] at (4.01, 0) {configuration \& input};
  \fill[aisiwarm] (8.30, -0.13) rectangle (8.64, 0.13);
  \draw[aisicrimson, dashed, line width=0.8pt] (8.30, -0.13) rectangle (8.64, 0.13);
  \node[anchor=west, text=aisinavy] at (8.76, 0) {custom / user-added};
\end{scope}
\end{tikzpicture}%
}
\caption{The \transect{} pipeline: \textbf{ingest}, \textbf{select}, \textbf{scan}, \textbf{store}, and \textbf{render}. The dataframes are used to generate the navigable report but can also be used for downstream cross-run analyses. Grey dotted boxes show inputs and configuration; the \emph{Spec} draws on evaluation materials and, where needed, expert input. \emph{Structural} scanners extract recorded information without model calls;  \emph{judged} scanners call the models selected for the analysis. Red dashed marks indicate two extension points. \emph{Custom scanners} add \emph{structural} or \emph{judged} signals to dataframes. \emph{Custom report layers} use those dataframes to add timeline layers or interface elements to the report without changing the upstream analysis.} 
\label{fig:transect_pipeline}
\end{figure}

\Cref{fig:transect_pipeline} shows the five stages. \textbf{Ingest} places the run on a shared \textbf{turn} axis, with one turn per agent output, so later signals use the same index across source formats. The axis records output order; the elapsed time between outputs can vary. \textbf{Select} chooses the sample and one or more epochs to analyse.\footnote{A \textit{sample} is a task instance with a specified prompt and evaluation setup; an \textit{epoch} is one attempt at that sample.} \textbf{Scan} runs the scanners. \textbf{Store} assembles their results into \transect{}'s dataframes. \textbf{Render} produces the report, linking scanner outputs to their source turns in the transcript. Users can export the returned dataframes and combine them across epochs for their own analyses.

A \textbf{scanner} extracts information or classifies behaviour from a transcript. In \Cref{fig:transect_pipeline}, \emph{structural} scanners extract recorded information, such as context compactions, operator messages, and per-turn token use, without calling a language model. \emph{Judged} scanners call the language models selected for the analysis to classify, rate, or annotate the evaluated agent's behaviour. The AI R\&D analysis below combines recorded operator messages and context compactions with research-activity labels from a custom judged scanner (\Cref{fig:tabpfn_timeline}).

The built-in \emph{judged} scanners run when judge models and the required Spec vocabularies are supplied. The \texttt{decision\_phases} scanner divides the run into contiguous phases and labels them using a \emph{closed, pre-registered vocabulary} based on the Spec, reusable across samples and epochs of the same family. The \texttt{subagent\_classification} scanner labels sub-agents from their delegation instructions: what the orchestrator asked them to do, which may differ from what they actually did. 

Users can add \emph{custom} \emph{structural} or \emph{judged} scanners to address further evaluation questions. For bespoke classification questions, the public \texttt{cohort\_llm\_scanner} helper gives a custom scanner the same judging machinery as the built-in scanners: voting across repeated judgments or multiple judges and a per-judgment provenance record identifying the judges, their individual answers, and any verifier changes. In solo and k-roll regimes, a verifier model can review selected classifications in an optional second round and revise their labels. Selection uses low-confidence or low-agreement triggers and random spot-checks. Nonetheless, these outputs remain fallible model judgments. Depending on the judging regime, they carry a model's stated confidence, within-model stability, or between-model agreement (\Cref{app:trust}). While stability and agreement help assess the reliability of a judgment or classification, they do not establish validity, as that would require comparison to a ground truth (usually, a human expert's validated judgement). The reviewer ultimately remains responsible for judging what the evidence supports.

\subsection{Design principles}
\label{sec:transect:principles}

Three design principles guide how the analysis is configured, interpreted, and extended. The first is to configure analysis for the evaluation \textbf{family}. Each family's \textbf{Spec} can define the behavioural vocabulary and any relevant sub-agent categories used to interpret its runs, drawing on evaluation materials and, where needed, subject matter expertise.

The second is structured expert elicitation  \citep{hanea2017idea, ohagan2006uncertain}. Users can ask a coding assistant, guided by the package's skills, to flag missing task context or unclear category definitions and help prepare questions for subject-matter experts (SMEs). Experts can then supply information missing from the evaluation materials, such as outcomes of interest, and review proposed classification rubrics. \transect{} records the category labels and definitions used by its built-in judged scanners. These records let users inspect the applied rubric and compare revisions across analyses.

The third is automation. The package's skills and worked examples guide coding assistants in running analyses and adding scanners or report views. They are designed to reduce the work needed to adapt the pipeline as evaluation tasks change. For a new family, users can supply a different Spec and add analyses while reusing the existing pipeline for supported transcript formats. The diagnostics skill helps users inspect disagreements, check the recorded judging setup and test sensitivity to the Spec's category names while keeping their definitions fixed.

A custom report layer draws on the same dataframes to add a new view (\Cref{app:package}). A design requirement is that report views preserve the available diagnostics and provenance of judged outputs. The report displays available confidence and agreement diagnostics and flags potential reliability problems. Provenance information helps reviewers understand the source of scanner judgements, and reliability information helps them assess what conclusions these judgements can support.

\subsection{Related work}
\label{sec:related}

\transect{} builds on Inspect AI for evaluation-log access and judge-model interfaces, and on Inspect Scout for transcript representation, scanner execution and stored results \citep{inspectai,dubois2026sevensteps,inspectscout}. Inspect Viz supplies the report's interactive charts.\footnote{\url{https://meridianlabs-ai.github.io/inspect_viz/}.} \transect{} organises these components around an evaluation family and enables co-interpretation of multiple scanner outputs on a shared timeline.

Docent is another prominent transcript analysis tool which uses language model assistants to help users interactively navigate through transcripts \citep{transluce_docent_plans}. \transect{}, in contrast, supplies a reusable analysis and reporting pipeline: users can apply a family Spec and their chosen judging settings across samples and epochs. The package's coding-assistant skills help users add analyses or views for new questions while reusing shared tables and report interfaces. Platforms such as LangSmith, Langfuse, Arize Phoenix, Weights \& Biases Weave, and AgentOps support trace inspection, debugging and application monitoring; LangSmith and Langfuse also support offline experiments and online evaluation.\footnote{\url{https://docs.langchain.com/langsmith/evaluation}; \url{https://langfuse.com/docs/evaluation/overview}; \url{https://arize.com/docs/phoenix}; \url{https://docs.wandb.ai/weave}; \url{https://docs.agentops.ai/}.} 

\transect{} brings family-defined activity labels, recorded events and token use together on an interactive turn-level timeline, so reviewers can examine how these signals relate within a run and follow them to the source transcript. \transect{}'s contribution is to combine adaptable analysis, interactive evidence review and inspectable model judgements in one reusable workflow for long-horizon evaluations. Methods for statistical analysis and evaluation design are complementary \citep{luettgau2025hibayes,pilditch2026optstop}, since \transect{} structures within-run evidence for such analysis but does not make sparse observations statistically sufficient.

%% file: sections/02b_Walkthrough_case.tex
\section{Case Study: Reading an AI R\&D Evaluation with Transect}
\label{sec:cases}

AI research and development (R\&D) is a prototypical example of an open-ended, long-horizon task for frontier agents. AI R\&D evaluations measure whether current agents can perform machine learning research that might accelerate AI development, including through so-called \textit{recursive self-improvement} \cite{wijk2024rebench,field2026views,anthropic2026buildsitself,chen2026rsi}. 
Such evaluations exemplify the difficulties of maintaining observability on agent behaviour. An agent may operate autonomously for days or weeks, generate billions of tokens, execute thousands of tool calls, and delegate work to hundreds of sub-agents \cite{oai-navier}. It is also unclear what successful behaviour looks like, since ``doing good science'' is a fuzzy construct \cite{bowkis2026automated,feyerabend1975against,laudan1983demise,polanyi1958personal}, and the final artefact is a written paper or report, whose quality may conceal many issues during its construction, including data fabrication, citation hallucination, and deep conceptual misunderstandings \cite{chen2025mlrbench,ansari2026compounddeception,bienz2026mysteriouscitations}.

We illustrate \transect{} on a single sample of the CRUX initiative's \emph{shadow evaluations} \citep{kirgis2026can}. In these evaluations, an agent receives a research brief based on a complete but unpublished human-written machine-learning research paper and is asked to produce a finished manuscript. A human operator is available over a Telegram chat interface to resolve operational problems but does not direct the research process itself.

The brief for this sample poses a methods problem in tabular prior-fitted networks (PFNs), transformers that classify tabular data through in-context learning over a support set rather than task-specific training. The research agent was asked whether a deployment-time detector could identify \emph{harmful} distribution shifts (changes that degrade the model's accuracy) without erroneously flagging benign shifts and without access to labels at inference time. The brief given to the agent requested a method to solve this problem which outperforms a set of pre-specified classical shift detector baselines and survives a pre-registered falsification test. The agent had to write and submit a manuscript that it had reviewed and deemed worthy of acceptance at a machine learning conference, along with a package for reproducing the analyses and results. 

The resulting OpenClaw transcript \citep{openclaw} sees a Claude-Opus-4.7 agent \cite{anthropic2026opus47card} produce some 554 orchestrator tool calls and 79 sub-agent deployments, receive 57 operator messages, and undergo 8 context compactions. 13 million tokens were generated during the evaluation, 89\% of them in delegated sub-agent work. 

Our purpose is not to provide a complete assessment of the agent's AI R\&D capability, for which we refer the reader to \citet{kirgis2026can}. Instead, we ask four questions a reviewer of this run would want answered: how the research process unfolded (\Cref{sec:cases:timeline}), where test-time compute was spent (\Cref{sec:cases:spend}), how delegated agents coordinated (\Cref{sec:cases:multiagent}), and what further questions can be asked by co-interpreting scanner outputs (\Cref{sec:cases:stats}). Each subsection reports an observation, discusses plausible interpretations, and notes how \transect{} produced it. \Cref{sec:cases:limits} then discusses what the case establishes and what it does not.

\subsection{How did the research process unfold?}
\label{sec:cases:timeline}

\begin{figure}[h]
\centering
\includegraphics[width=\textwidth]{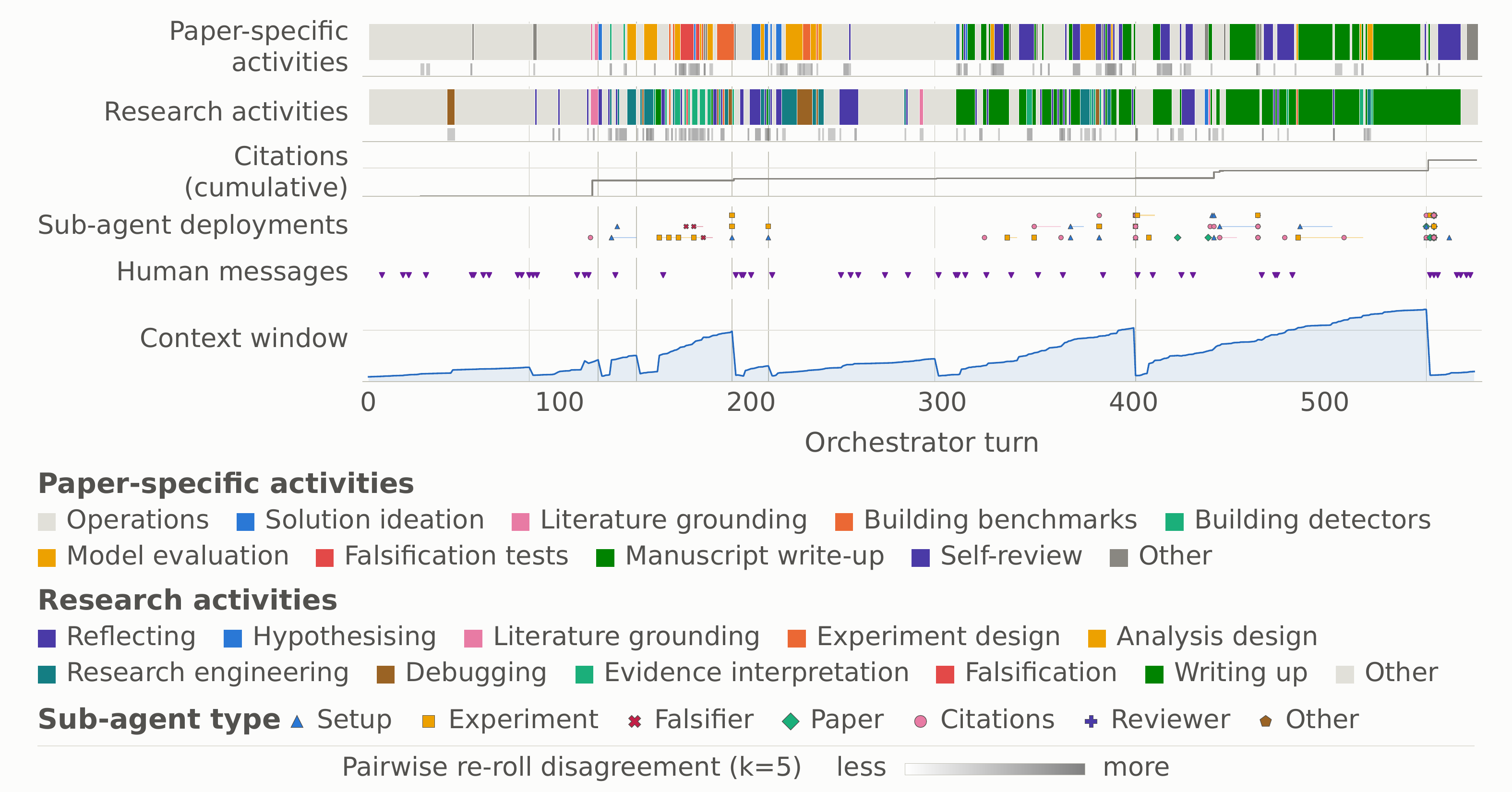}
\caption{The TabPFN epoch on an orchestrator-turn axis: paper-specific activities, research activities, cumulative references, sub-agent deployments, operator messages, and context-window size. Grey strips show pairwise label disagreement across five judge rolls for every turn; darker means greater disagreement.}
\label{fig:tabpfn_timeline}
\end{figure}

\Cref{fig:tabpfn_timeline} places six layers on a shared \textbf{orchestrator-turn} axis, where one turn corresponds to one orchestrator output, so that the layers can be read against one another. The package's shared model-turn axis assigns an index to each recorded orchestrator model output (\Cref{app:scanners}). The lower layers come from \emph{structural} scanners, which read facts already recorded in the transcript without calling another model (in ascending order): how the orchestrator's context changed across compactions, when the operator sent a Telegram message, when sub-agents were launched and returned, and, from a custom scanner that identifies arXiv identifiers, DOIs and heuristically matched titles, where references first appear. Sub-agent markers are labelled using the built-in \texttt{subagent\_classification} scanner, which describes what the orchestrator asked each sub-agent to do, not necessarily the work it ultimately performed. The two upper bands come from custom \emph{judged} scanners that assign one principal label, or a residual ``Other'', to each of the epoch's 580 orchestrator outputs, including tool-only outputs. The research-activity band uses ten named categories intended to capture the typical structure of AI R\&D tasks: \textit{Hypothesising}, \textit{Literature grounding}, \textit{Experiment design}, \textit{Analysis design}, \textit{Falsification}, \textit{Research engineering}, \textit{Debugging}, \textit{Evidence interpretation}, \textit{Reflecting} and \textit{Writing up}. The paper-specific band uses the closed vocabulary defined for \texttt{decision\_phases} in the family's \textbf{Spec}, which follows the task's concrete research questions and deliverables, from \textit{PFN problem framing} and \textit{PFN solution ideation} through \textit{Benchmark construction}, \textit{PFN model evaluation} and \textit{Falsification tests} to \textit{Manuscript write-up}, \textit{Self-review} and \textit{Code reproducibility}, and separately tracks \textit{Non-research-relevant engineering}. The category definitions are given in \Cref{app:calibration}.

The value of the timeline comes from reading these independently produced layers together. The epoch begins with a long operations-dominated stretch containing little judged research activity other than debugging. This also coincides with frequent operator messages, mostly helping the model to establish permissions for its various tools. Inspection of these turns shows the agent establishing the environment's permissions, resources, and available tools. Much of the initial literature engagement follows this operational period. The middle of the epoch alternates between benchmark construction, model evaluation, and falsification, with several delegated sub-agents supporting those activities. The final portion concentrates on manuscript drafting and self-review, with writing, citation, and reviewer sub-agents used to parallelise that work. Some context compactions also occur near sub-agent deployments, exposing a possible relationship that can be further investigated rather than merely noticed while reading the raw transcript.

The combined view also makes absences salient. In the current analysis, only two of the 580 orchestrator outputs are labelled \textit{Hypothesising}, compared with 56 labelled \textit{Reflecting} and 27 labelled \textit{Evidence interpretation}. Surprisingly, \textit{Analysis design} does not appear at all. The two hypothesising labels occur when the orchestrator plans to derive novel theoretical results for a detector it has already proposed, and it delegates this work to a sub-agent both times. This pattern has at least two potential explanations: the agent may have performed little hypothesis formation, or the scanner may under-detect diffuse or secondary activities when assigning a single principal label. A missing label is therefore weak evidence that the behaviour was absent. Even with that limitation, the timeline turns the impression that ``the agent rarely seems to ideate'' into a reproducible observation linked to specific turns, an explicit rubric, a reliability indicator, and competing explanations for review.

Both activity bands were produced in a \textbf{$k$-roll} regime: Claude Opus 4.7 classified each output $k=5$ times, reading the target's full text and tool arguments with up to three preceding and following outputs, and the bands display the modal label with adjacent identical labels merged, forming 129 contiguous phases for the paper-specific activities. The grey strip beneath each band shows the fraction of the ten judge-roll pairs assigning different labels: 474 of 580 outputs (81.7\%) receive unanimous research-activity labels and 476 (82.1\%) unanimous paper-specific labels, with \textit{Reflecting} and \textit{Evidence interpretation} comparatively unstable. A \textbf{cohort} regime could instead call several judge models to expose between-model agreement. A \textbf{solo} run would call one model once and would therefore be cheaper, but would provide no empirical stability estimate. The stored scan results retain the prompts, inputs and model responses behind every label so that they can be inspected transparently. Variation in judgements helps identify labels that merit caution or rubric refinement, but low variation does not establish that the labels are \textit{valid}, which would require independently collected human labels or another suitable external referent to calculate false positive and negative rates (\Cref{app:trust}).

\subsection{Where was test-time compute spent?}
\label{sec:cases:spend}

\begin{figure}[h]
\centering
\includegraphics[width=\textwidth]{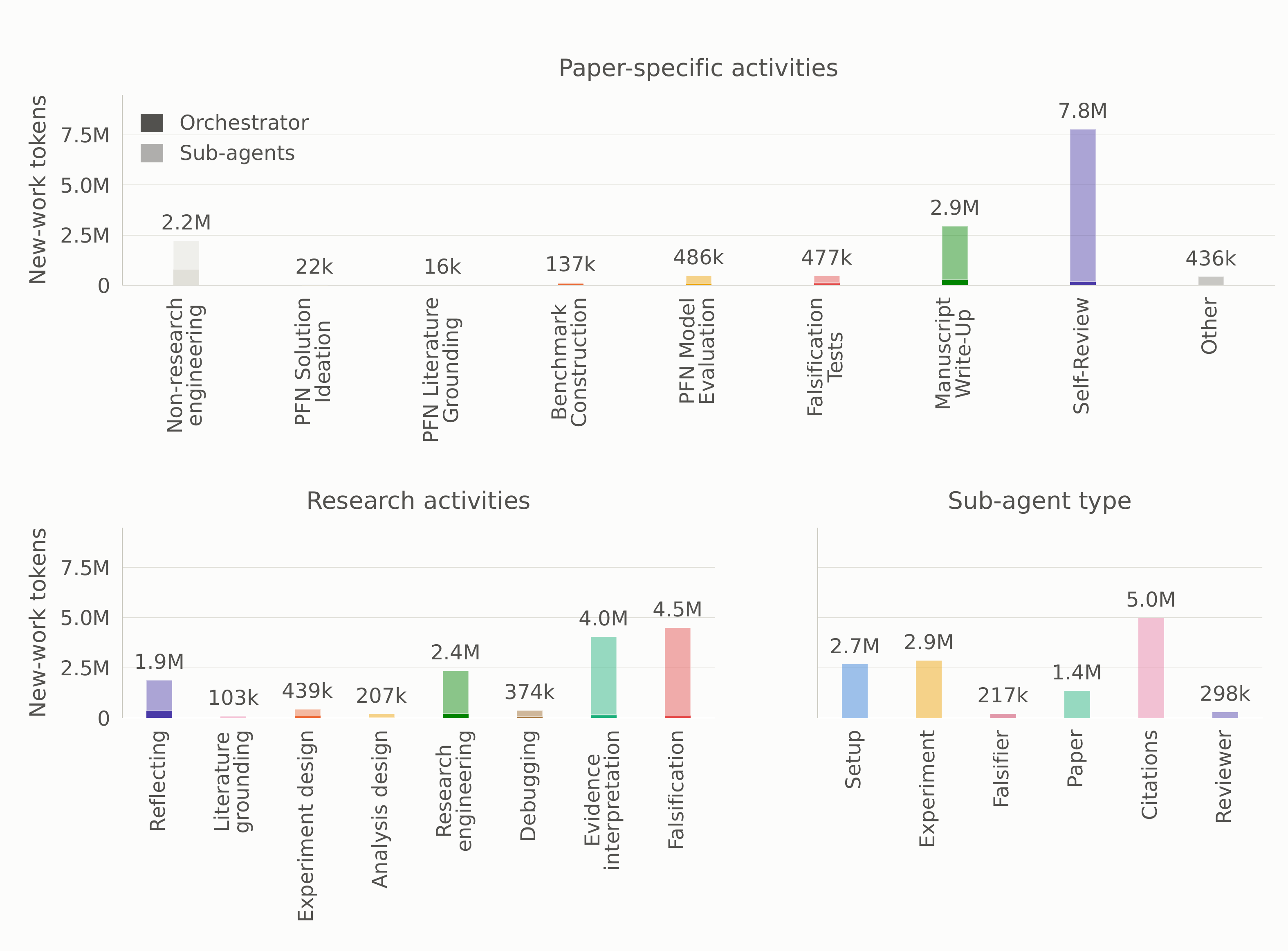}
\caption{New-work tokens attributed to paper-specific activities, research activities, and sub-agent categories. In the activity panels, solid segments show the orchestrator's own expenditure and lighter segments the expenditure of sub-agents launched within the unit. The research activities panel excludes work without an assigned research skill, and the sub-agent panel excludes the orchestrator's own activity. Token counts should not be summed across panels.}
\label{fig:tabpfn_spend}
\end{figure}

The same turn-level decomposition supports a second question: how was the test-time compute allocated? Token telemetry, activity labels, and sub-agent events share the same segmentation, so token expenditure can be attributed to the activity and the sub-agent responsible for it. \Cref{fig:tabpfn_spend} shows the three attributions side by side.

In the current analysis, self-review and manuscript write-up account for roughly three quarters of all new-work tokens: 7.8 million and 2.9 million, respectively, out of 14.5 million. Almost all of this expenditure is delegated, reflecting the orchestrator's use of citation, writing, and reviewer sub-agents during the final stages. Non-research engineering is the third largest category at approximately 2.2 million tokens, much of it associated with resolving permissions and determining the scope of the evaluation environment.

By contrast, solution ideation only accounts for approximately 22,000 tokens. Some relevant work may have been labeled under reflection, research engineering, or evidence interpretation, which account for substantially more expenditure. Nevertheless, this allocation is consistent with the original evaluation's observation that agents ``produced papers with extremely thorough negative findings rather than papers with new ideas'' \citep[p.~16]{kirgis2026can}. Citation sub-agents consume approximately 5 million tokens, but much of this work occurs during manuscript drafting and self-review rather than early hypothesis development. One plausible reading is therefore that literature search primarily supports polishing and substantiating the manuscript rather than generating new scientific directions.

These findings also bear on harness and setup uncertainty. Non-research operations occupy a large fraction of the epoch, and most of the operator messages occur during those phases, often providing permissions, resources, or credentials needed to overcome an infrastructural impasse. The observed allocation could reflect an agent preference for tractable operational and writing work, a scaffold that inadequately supports scientific work, or scientific reasoning that is not exposed in the recorded transcript. \transect{} cannot distinguish these explanations by itself. It makes them precise enough, however, to motivate interventions: evaluators could change the initial permissions, provide a different tool set, alter the research brief, or add progress gates, and then compare the resulting phase and token distributions across epochs.

Token allocation also provides a way to study capability as a function of test-time compute. Two systems may produce similarly rated final artefacts while using compute differently: one may brute-force many weak experiments, while another may perform fewer, better-motivated tests. Likewise, apparent improvement may be concentrated in research engineering or manuscript production rather than in the scientific activities relevant to capability forecasts, such as solution ideation, experiment design, or falsification. One transcript cannot establish these differences, but the exported dataframes make the same decompositions available across agents, budgets, scaffolds, and repeated epochs.

\subsection{How did delegated agents coordinate?}
\label{sec:cases:multiagent}

\begin{figure}[h]
\centering
\includegraphics[width=\textwidth]{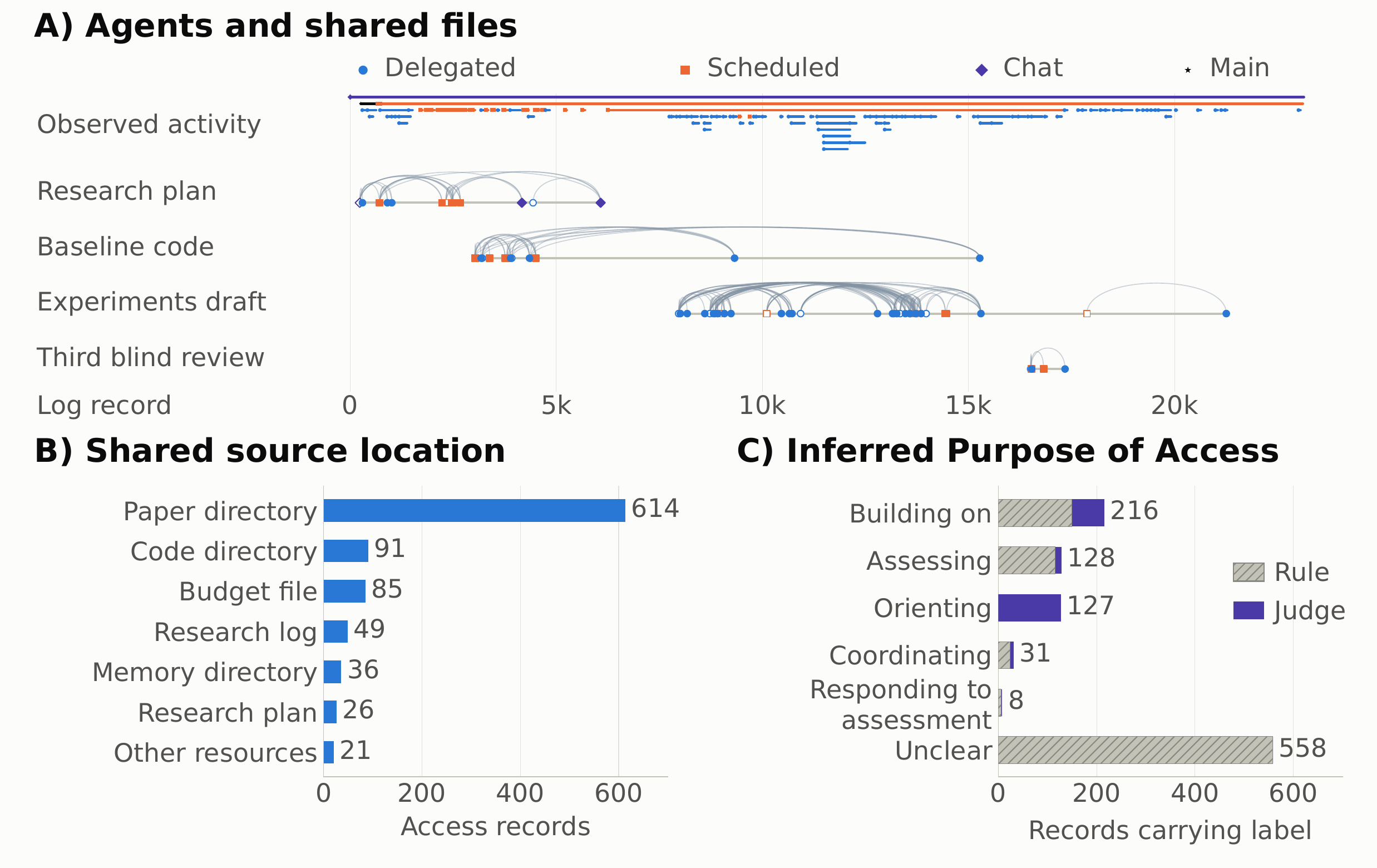}
\caption{Multi-agent analysis of one TabPFN epoch. A: activity across 131 agent sessions and four shared-file tracks in source-log order. Arcs connect earlier writes (open symbols) to later cross-agent reads (filled symbols); colour and shape denote context type. B: all 922 access records by resource path. C: inferred purposes, with compound labels counted in each category. Access does not establish uptake.}
\label{fig:tabpfn_multiagent}
\end{figure}

A multi-agent evaluation creates an interaction structure that is difficult to reconstruct from raw logs: which shared files agents wrote and read, in what order, and what kinds of exchange these were. To follow work across agent boundaries we extend the report with a custom \emph{structural} scanner for shared-file access and an optional \emph{judged} scanner for its likely purpose, and combine their outputs in \Cref{fig:tabpfn_multiagent}.

The structural scanner joins recorded file access events to earlier writes by other separately logged agent conversations or tasks (\emph{sessions}), preserving session identifiers, resource paths, source positions, and evidence types in the \textbf{dataframes}. It produces 922 access records across 131 participating sessions, of which 346 records connect one delegated sub-agent to another. The sessions comprise delegated sub-agents, tasks triggered by the harness's scheduler, and the harness's main and Telegram chat contexts. The dataframes retain the writer-reader direction.

\Cref{fig:tabpfn_multiagent}, Panel A follows the research plan, baseline code, experiments-section draft, and third blind-review document through the recorded workflow. These four tracks represent 159 access records involving 39 sessions; the activity strip retains all 131 participants. Each arc connects a record's first qualifying read to the named writer's latest preceding write. Agent bars span first to last recorded activity, while file tracks span first recorded write to last recorded read or write. The axis follows source-file order: these spans establish neither elapsed time nor continuous agent activity or file persistence.

Access is concentrated in the manuscript workspace: 614 of the 922 records (66.6\%) concern paths under \texttt{paper/}, including drafts, reviews, and build files. The experiments-section draft alone accounts for 102 records, visible as repeated connections along its track. Together with the token attribution in \Cref{sec:cases:spend}, this identifies the manuscript workspace as a priority for examining how delegated contributions were assembled. A reviewer can follow a shared document to its recorded writers and readers, then inspect their instructions and subsequent edits.

Purpose classification helps reviewers follow concrete exchanges. For example, the transcript records a sub-agent opening the third blind-review document after receiving instructions to address five major and four minor presentation findings and rebuild the manuscript. All five judge rolls classify this access as \emph{responding to assessment}. This connects the review track in panel A to a revision brief and a recorded read, giving the reviewer a concrete trail to investigate. Inspecting subsequent edits would establish which findings were addressed and how.

This remains a prototype, but its provenance, coverage, and repeatability diagnostics make the limits explicit. Across the transcript, 150 records receive substantive rule-assigned labels and 214 receive judge-assigned labels; 558 remain unclear. Among the judge-labelled records, 166 (77.6\%) receive unanimous labels across five Opus 4.7 rolls. This measures within-model repeatability; purpose validity and uptake require further evidence. The coverage gaps suggest concrete improvements: parsing omitted shell commands and direct messages would capture more exchanges, while harness records of read results, file versions, and sub-agent return payloads would help reviewers trace contributions and check the inferred purposes.

\subsection{What else can we learn about AI R\&D?}
\label{sec:cases:stats}

\begin{figure}[h]
\centering
\includegraphics[width=\textwidth]{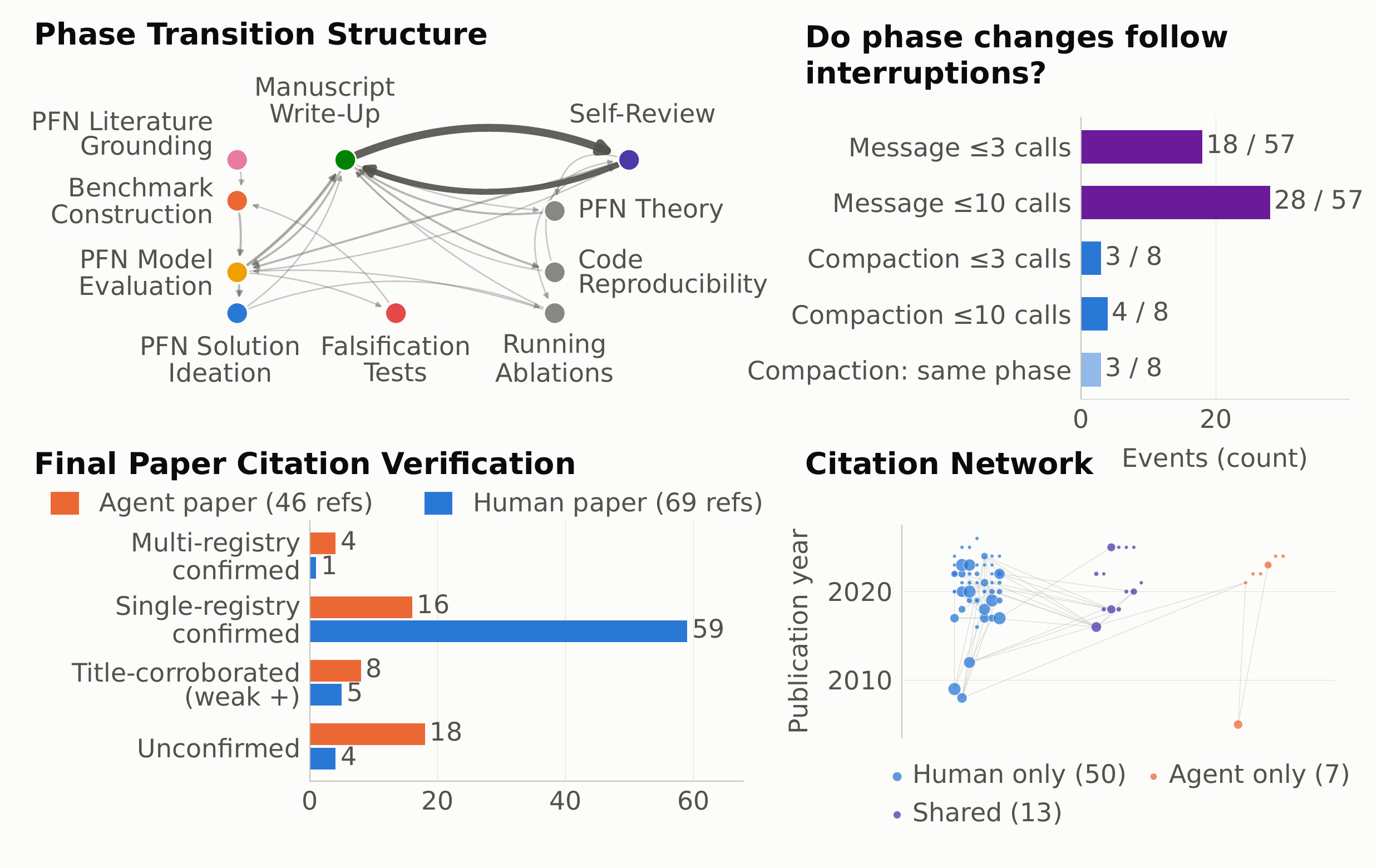}
\caption{Descriptive analyses of the TabPFN epoch. Top left: observed transitions between paper-specific phases after removing non-research-relevant engineering phases and merging consecutive repeats, with edge width proportional to transition count. Top right: the number of operator messages and context compactions followed by the start of a new phase within 3 or 10 orchestrator tool calls, and compactions after which the current phase continues; these are co-occurrence counts, not effects. Bottom left: registry verification of references in the agent- and human-written manuscripts against OpenAlex \citep{priem2022openalex}, Crossref \citep{crossref_docs}, and arXiv \citep{arxiv}. Bottom right: the combined citation network, with node size indicating global citation count and edges indicating citations among works in the combined bibliography.}
\label{fig:tabpfn_stats}
\end{figure}

Once phase labels, recorded events and reference records share one representation, questions that would otherwise require bespoke transcript processing become cheap to formulate and test. We ask three here: whether the workflow has recurring structure, whether behavioural transitions cluster near operator messages or context compactions, and how the agent's literature coverage compares with that of the human authors. \Cref{fig:tabpfn_stats} presents descriptive analyses of these three questions.

The phase-transition graph shows the transitions between paper-specific research activities, excluding non-research engineering. Two regimes are visible: a build-and-test loop through benchmark construction, model evaluation, ablations, and falsification; and a write-and-review oscillation between manuscript write-up and self-review, with some input from theory building, code reproducibility, and running ablations.

This phase-transition graph  (top left panel, \Cref{fig:tabpfn_stats}) illustrates how an ordered sequence of task-specific phase labels makes transitions between activities directly computable.

This graph structure can also guide investigation of the evaluation setup. The repeated cycle between manuscript drafting and review motivates testing whether the setup encourages iteration over the written artefact. The build-and-test loop is less dominant and does not show an equivalently strong earlier cycle of generating, selecting, and testing alternative ideas. Evaluation and harness designers could use this graph as a target for intervention, for example by introducing an explicit ideation gate or requiring comparison among several candidate methods before manuscript drafting begins. Comparing repeated epochs under otherwise comparable original and modified setups would help assess the intervention's effects on behaviour and the final score.

The event-proximity panel (top-right panel, \Cref{fig:tabpfn_stats}) shows whether new phases begin near operator interventions or context compactions. In the current analysis, a new phase begins within three indexed steps of 18 of the 57 operator messages and within ten steps of 28 operator messages. New phases begin within 3 indexed steps of 3 compactions and within 10 steps of one further compaction. These are raw co-occurrence counts, not causal effects: operator messages may respond to an already emerging transition, and compactions may occur at points where the agent would have changed direction anyway. Nevertheless, both event streams are partly determined by the evaluation setup, so their proximity to behavioural transitions identifies a concrete source of setup uncertainty for controlled follow-up.

The final two panels use a feature of the shadow-evaluation design: the agent-written manuscript can be compared with the unpublished human-written paper on which the brief was based. We first verify cited references against OpenAlex \cite{priem2022openalex}, Crossref \cite{crossref_docs} and arXiv \cite{arxiv}. 20 out of the agent's 46 cited references are verified in at least one of these registries, and a further 8 have title matches without linked DOIs, leaving 18 unconfirmed. In contrast, 60 out of the human paper's 69 references are verified in at least one registry, with a further 5 having a title match, leaving only 4 unconfirmed references. Registry non-matches do not prove fabrication, but the difference identifies a substantially greater citation-verification burden in the agent-written manuscript.

The citation network provides a complementary view. The manuscripts share 13 references, but the human paper draws on a larger and more internally connected literature, as indicated by the density of citation links among its referenced works. This is consistent with the possibility that the agent's search did not recover the same scientific background as the human researchers. It also suggests a follow-up intervention: evaluators could supply the human bibliography as part of the scaffold and test whether access to the same literature changes the agent's hypotheses, experimental choices, or final claims.

The package supports displaying analyses such as these, alongside its timeline and token allocations, as custom report elements (\Cref{fig:transect_pipeline}, \Cref{app:enrichment}). Linking their summaries to phases and source turns requires matching transcript identities and compatible turn or tool-call indices.

\subsection{What this case establishes}
\label{sec:cases:limits}

This analysis establishes some interesting features of the agent's behaviour: It starts with a long operational opening without much hypothesis generation. Most of the tokens are spent by sub-agents doing manuscript writing and review, phases through which the agent iterates several times.

However, this analysis is descriptive rather than confirmatory. It does not firmly establish the agent's AI R\&D capability, the validity of the judged labels, or causal effects of operator messages, compactions, or the scaffold. Those claims would require multiple epochs on multiple samples from the same evaluation family, independently collected expert labels for the judged scanners, and an explicit statistical model for trends across runs (\Cref{app:datamodel}). \transect{} offers a coherent framework for doing this kind of rigorous, confirmatory analysis on open-ended, long-horizon evaluations, while maintaining reproducibility and transparency.

%% file: sections/03_conclusion.tex
\section{Conclusion}
\label{sec:conclusion}

Open-ended, long-horizon agent evaluations are narrowing the observability envelope---what an evaluation reliably tells us about an agent---because they are getting longer, more complex, and more parallelised. A natural solution to this is to use language models to assist with transcript analysis, but this leads to significant researcher degrees of freedom. We present \transect{} as a tool to address these twin problems, by making within-run behaviour scrutable while supporting reliability checks and reproducible analysis. \transect{} places a run on a shared turn axis, selects the sample and epoch to read, reads the transcript with structural and judged scanners, and writes every signal into one set of dataframes that can be digested as a human-readable report. This is the heart of the value proposition of \transect{}: it allows evaluators to both readily access the `wider picture' and quickly identify and analyse key moments in the agent's trajectory - whether these are indicative of agent capability, safety incidents, harness constraints, or other research questions the decomposition exposes. We argue that this human-accessible, cohesive picture is critical for maintaining effective oversight for increasingly long and complex evaluation transcripts.

While language models are becoming increasingly important for transcript analysis, they must be used with caution. LLM-based scanners can overstate confidence, hide uncertainty, and ultimately be over-interpreted by human reviewers \cite{bavaresco2025llms,romeo2026exploring}. Language models are increasingly capable as judges for many classification tasks, though reliability does not establish validity \citep{norman2026reliability}. \transect{} records available reliability diagnostics and flags potential problems for review. Assessing validity requires separate evidence, such as comparison with independently collected expert annotations.

Expert elicitation, long established in risk analysis \citep{hanea2017idea}, can calibrate \transect{} through written materials or targeted questions that shape which signals a family of evaluations should surface. Reviewers can examine classifications alongside the agent's recorded reasoning, tool use, and harness events. These signals need not be independent, but reading them together can expose gaps or inconsistencies that warrant inspection. Provenance records show how a classification was produced and which source turns it was derived from. This view can also help evaluators investigate how the harness shapes agent behaviour and where its records are incomplete. For example, if the transcript does not record a tool failure, adding that event to future logs would let reviewers examine subsequent behaviour alongside the failure. The package's coding assistant skills and worked examples are designed to reduce the work needed to configure new evaluation families and extend their analyses. They guide coding assistants in adding scanners or report views, inspecting judging records and diagnostics, and testing sensitivity to category names. These provenance and reliability disciplines support reproducibility and should accompany each extension of the analysis.

Transect turns long evaluation transcripts into source-linked reports and reusable analysis tables. The case study illustrates how these outputs bring behavioural classifications, token use across the orchestrator and sub-agents, multi-agent interactions, and harness events into one review workflow that is navigable, auditable, and reproducible.

\section*{Acknowledgements}

We are grateful to the authors of the CRUX shadow evaluations \citep{kirgis2026can} for providing the data for the case study presented in this paper. We would also like to thank Peter Kirgis, Sayash Kapoor, Andrew Schwartz, Stephan Rabanser, Viet Nguyen, Arvind Narayanan, Michael Schmatz, Giorgi Giglemiani, Jessica Wang, Pawan Mudigonda, Magda Dubois, Auss Abbood, Orazio Angelini, and Ole Jorgensen for thoughtful discussion throughout the development of \transect{}. Finally, we extend our thanks to Meridian Labs \citep{meridianlabs} for their close collaboration in the development of the package.

\newpage

%% file: sections/04_appendix.tex


\newpage

\section{The Transect Package}
\label{app:package}
The descriptions below apply to Transect package version 0.2.0.
This appendix describes how to run an analysis (\Cref{app:pipeline}), what the scanners produce (\Cref{app:scanners}), and how to read and reuse the outputs (\Cref{app:datamodel}). It then covers family configuration (\Cref{app:calibration}), judging and auditing (\Cref{app:trust}), extensions (\Cref{app:enrichment}), and reproducibility and scaling (\Cref{app:reproducibility}).

\subsection{Running an analysis}
\label{app:pipeline}

To analyse a run, the user supplies transcript logs and a \textbf{Spec}, the reusable configuration for an evaluation family. The Spec defines the behavioural vocabularies and task context. Analysis settings, including the judge models and number of repeated calls, are supplied separately to \texttt{transect()}. This function analyses the selected sample and epochs, generates the report, and returns a \texttt{TransectResults} object containing the dataframes.

\texttt{load()} rebuilds dataframes from a stored scan without model calls. \texttt{render()} generates a report from those results and available source transcripts (\Cref{app:datamodel}).

\transect{} builds on Inspect Scout's transcript representation, scanner abstraction and stored scan results \citep{inspectscout}.

\Cref{tab:core} summarises the pipeline's inputs and outputs.

\begin{table}[h]
\centering
\small
\begin{tabularx}{\linewidth}{@{}l X X@{}}
\toprule
\textbf{Stage} & \textbf{Input} & \textbf{Output} \\
\midrule
Ingest & Logs in a supported format & Scout transcripts representing agent and sub-agent activity recoverable from the source \\
Select & Transcripts and sample/epoch settings & The selected sample and one or more epochs \\
Scan & Selected transcripts, Spec and analysis settings & Results from structural scanners and any enabled judged scanners, saved in a Scout scan store \\
Store & Stored scanner results & Tables of activities, recorded events, token use and audit information \\
Render & Dataframes and source transcripts & HTML reports organised by epoch; tables users can analyse or export \\
\bottomrule
\end{tabularx}
\caption{Inputs and outputs of the five pipeline stages. Structural scanners extract or derive information without model calls; judged scanners call language models. Regenerating a report with excerpts requires access to the source transcripts.}
\label{tab:core}
\end{table}

\paragraph{Ingest.}
Ingestion presents supported evaluation logs as Scout transcripts so the same scanners can analyse them. \transect{} reads Inspect \evalfile{} logs and imports OpenClaw JSONL exports produced by telemetry-hal. The OpenClaw importer removes duplicate records from cumulative exports and reconstructs recorded messages, tool activity and sub-agent activity. What reviewers can inspect depends on the source records, including the available event and token-usage details. Additional formats require an adapter to this representation (\Cref{fig:transect_pipeline}).

\paragraph{Select.}
A sample is an evaluation task or item; an epoch is an attempt at that sample. Each analysis selects one sample and one, several or all of its epochs before scanning. Inputs containing multiple samples require an explicit sample choice. If no epochs are specified, \transect{} selects the earliest epoch recorded as successful, or the earliest epoch if none is recorded as successful. This default favours successful attempts when available. Analyses requiring all attempts or a pre-specified sample of attempts should set the epoch selection explicitly.

When several epochs are selected together, the same configured scanners are applied to each attempt. Once scanning finishes, \transect{} builds the dataframes and generates separate reports for the selected epochs. The built-in dataframes retain results from all selected epochs; users can select an attempt by filtering the \texttt{epoch} column. Transcript identifiers link the results to their attempts, supporting separate review and comparisons across attempts.

\paragraph{Defaults.}
\Cref{tab:defaults} lists selected defaults and text limits in the reviewed package. The Spec supplies vocabularies and context. These defaults have not been established as optimal settings.

\begin{table}[h]
\centering
\small
\begin{tabularx}{\linewidth}{@{}X X l@{}}
\toprule
\textbf{Parameter} & \textbf{Default or limit} & \textbf{Control} \\
\midrule
Judge models & None; built-in judged scanners disabled & Top level \\
Roll count $k$ & 1 & Top level \\
Verifier & On for solo and k-roll; unavailable for cohort & Top level \\
Verifier model & First configured judge model & Top level \\
Verifier spot-check share & 5\%; default phase-sampling floor: 3 phases & Top level \\
Phase digests per classification chunk & 40 & Scanner \\
Phases per verifier chunk & 8 & Scanner \\
Phase narration & On & Scanner \\
Activity-informed sub-agent classification & Off & Scanner \\
Phase message excerpt, reasoning excerpt and each delegation string & 300 characters each & Scanner \\
Extracted task-prompt context for phases & 2{,}400 characters & Helper \\
Sub-agent task text & 800 characters & Fixed \\
Phase fill confidence & 0.3 & Fixed \\
Review trigger: classifier confidence or agreement & Below 0.6 & Fixed \\
Label-change gate: verifier confidence & At least 0.6 & Fixed \\
Response caching & On; later rolls use separate cache scopes & See text \\
\bottomrule
\end{tabularx}
\caption{Selected defaults in the reviewed package. Top-level controls are analysis-function arguments; scanner and helper controls require calls to the corresponding lower-level functions. Fixed values are set in the implementation. None of these controls is a built-in Spec field.}
\label{tab:defaults}
\end{table}

The top-level controls are \texttt{judge\_models}, \texttt{k\_rolls}, \texttt{verify}, \texttt{verifier\_model} and \texttt{verify\_sample}. Phase and sub-agent spot-checking use different selection rules, described in \Cref{app:trust}.

The phase scanner exposes \texttt{chunk}, \texttt{snippet\_chars}, \texttt{verify\_chunk} and \texttt{narrate}; the sub-agent scanner exposes \texttt{activity}. The task-prompt extraction helper accepts a \texttt{cap} argument; the built-in phase scanner uses its default. A digest may contain several delegation strings, so the 300-character limit does not bound its total length. Response-cache controls are available on \texttt{decision\_phases} and \texttt{cohort\_llm\_scanner}, but are not exposed by the top-level analysis function or the built-in sub-agent scanner (\Cref{app:reproducibility}).

This table is not intended as a complete record of the configuration needed to reproduce a run.

\subsection{What the scanners produce}
\label{app:scanners}

\paragraph{The turn axis.}
The report aligns behavioural phases, token use and recorded events on a shared \textbf{model-turn} axis. A turn is a model event with output; these events are numbered consecutively in normalised transcript order. Recorded sub-agent model turns use the same index, with separate lanes identifying the agents. Human-input markers use the same axis. An intervention recorded as a message is placed at the first model turn whose recorded input includes it; if that link is unavailable, its position is inferred from the recorded message and event order. Reconstruction may place a sub-agent's activity after the event that spawned it. Turn positions alone do not establish elapsed time, exact ordering across agents or whether agents acted concurrently.

\paragraph{Scan: two kinds of scanner.}
A scanner extracts information or detects patterns in a transcript \citep{dubois2026sevensteps,inspectscout}. \transect{} combines \emph{structural} scanners, which extract recorded information without model calls, with \emph{judged} scanners, which use language models to classify agent behaviour.

The built-in structural scanners are:
\begin{itemize}
    \item \texttt{token\_timeline} records the token usage reported for each model turn, its tool-call count and its agent lane. It also extracts sub-agent tool activity for placement on the timeline.
    \item \texttt{context\_flush} extracts recorded context-compaction events and their available token counts before and after compaction. For Inspect logs, it also retains the summarisation call's formatted prompt and the pre-compaction memory warning verbatim where recoverable. The report displays these texts under Eval setup / Core setup.
    \item \texttt{human\_intervention} extracts mid-run messages marked as operator or human input, recorded answers to agent-initiated requests for input, and tool-approval decisions recorded by Inspect's built-in human approver. Detection uses source annotations and event types.
    \item \texttt{eval\_setup} extracts initial system and task prompts, agent and task settings, and evaluation-log metadata where available. These records describe the conditions under which the agent ran.
\end{itemize}
The dataframe builders use these records to calculate token measures. They can also infer compaction candidates from drops in context size, keeping inferred candidates distinguishable from recorded events (\Cref{app:tokens}).

The built-in \emph{judged} scanners require models supplied through \texttt{judge\_models} and matching vocabularies in the Spec. \texttt{decision\_phases} uses the phase vocabulary; \texttt{subagent\_classification} uses the sub-agent categories. Both use shared judging machinery with solo, repeated-call and multi-model modes. Their labels remain fallible model judgements. Voting, optional verification and the confidence or agreement information available in each mode are described in \Cref{app:trust}.

\paragraph{\texttt{decision\_phases}.}
The scanner divides the run into contiguous phases labelled from a closed vocabulary comprising the Spec's categories and operational or fallback categories added where needed. An operational category covers setup and coordination; users can designate their own category for this purpose. The scanner adds \texttt{none\_of\_the\_above} if it is absent from the Spec, allowing judges to flag activity that does not fit the supplied vocabulary. Providing this option is good classification practice: forcing a choice among inappropriate categories can produce consistent but incorrect labels. Judges can still overlook a mismatch. The vocabulary should be fixed before analysing runs intended for comparison (\Cref{app:calibration}). 

Judges receive length-limited digests of main-agent message text, available recorded reasoning and delegation information, together with tool names. Reasoning excerpts are marked \texttt{[THINKING]}; a turn containing only readable recorded reasoning is also eligible. For redacted or summary-only reasoning blocks, the digest uses any readable summary supplied by the source. Recognised OpenClaw failure placeholders are excluded from the digests. The judges propose phase boundaries and labels for batches of digests, with the preceding phase label supplied as context where available. The results are combined and adjacent phases with the same label are merged, then mapped onto the shared turn axis. Some turns are assigned phase labels without their content being directly judged; the tables record the basis of each turn's label (\Cref{app:datamodel}). A continuous phase band therefore does not mean every turn was independently classified. 

When enabled, narration runs after classification and any verification. The first configured judge model receives phase labels, a limited selection of each phase's digests, and available task and Spec context. It produces a headline, short summary and titled groups of consecutive turns for the phase cards. If no usable narrative is returned for a phase, a label-based headline and empty summary are used instead.

Agreement on phase labels does not assess the accuracy of this generated prose. The classifications describe activities; conclusions about competence, correctness or safety require further assessment (\Cref{app:trust}).

\paragraph{\texttt{subagent\_classification}.}
This scanner assigns one label to each sub-agent span it classifies (a recorded sub-agent execution), using the Spec's sub-agent categories and \texttt{none\_of\_the\_above} for work that does not fit them. In the default \emph{instruction-based} mode, the judge reads the delegation instructions and labels the work requested. If a sub-agent is asked to survey literature but spends its run debugging, the judge still assesses the request.

In the experimental \emph{activity-informed} mode, the judge also receives a limited digest of recorded sub-agent activity, including message excerpts and tool-use counts where available. The prompt directs it to prioritise recorded work when this differs from the request. Both modes require task text recovered from delegation instructions or handoff messages. When activity records are absent, the activity-informed mode also relies on that text. Each classification covers the span as a whole, without dividing the sub-agent's work into behavioural phases.

\paragraph{Token measures.}
\label{app:tokens}
\texttt{token\_timeline} retains the reported input, output, total, cache-read, cache-write and reasoning token fields. Cache reads and writes count tokens retrieved from or stored in the prompt cache. The dataframe builder assumes Inspect's ModelUsage convention: input tokens exclude cache reads and writes, and output tokens include reasoning tokens. Importers using other conventions must normalise their counts before dataframe construction. The OpenClaw importer uses a fixed mapping to these fields.

The package then calculates four measures for each turn:
\begin{itemize}
    \item \texttt{context}: input plus cache writes and reads, representing the reconstructed input-context size.
    \item \texttt{new\_work}: input and output, plus cache writes capped at the non-negative increase in context since the previous turn with usable token records in the same agent lane.
    \item \texttt{billable}: input, output, and all cache-write tokens. Cache reads are excluded.
    \item \texttt{turn\_total}: input, output, cache writes and cache reads. This derived total can differ from the source's reported \texttt{total\_tokens}.
\end{itemize}
A missing input-token count or an all-zero input/output/cache record leaves these derived measures missing; missing components count as zero for this check. The check does not use reasoning or reported total tokens. Other missing components are also treated as zero when computing a derived value. The preceding context starts at zero in each lane and is preserved across skipped turns.

\paragraph{Interpreting token totals.}
The \texttt{billable} field is an unweighted token sum, not a monetary cost. It excludes cache reads even when a provider charges for them; prices can also differ for input, output and cache writes. Grouping token use by behavioural labels shows how tokens are distributed across the labelled activities; token expenditure alone does not measure scientific effort, research quality or efficiency.

\subsection{Reading and reusing outputs}
\label{app:datamodel}

\paragraph{Report.}
In the default layout, the report opens with information about the transcript and evaluation setup, drawn from the available source records. Its timeline aligns available behavioural phases, token use, sub-agent activity and markers for human interventions and recorded or inferred context compactions. Reviewers can inspect these signals together and open phase cards containing labels, generated narratives and expandable transcript excerpts. Where the source provides readable reasoning or a reasoning summary, these excerpts display it in a muted \texttt{[thinking]} line above the turn's text. The cards and excerpts can be explored within the generated report without Scout. Optional \emph{Open in Scout Viewer} links support further exploration of the full transcript through a running Scout viewer. When generating a report, users can configure section order and add custom report elements based on their own scanners or analyses (\Cref{app:enrichment}).

The reliability audit describes how the included judged classifications were produced and presents the available confidence, agreement and verification diagnostics (\Cref{app:trust}). Reviewers can use these details to investigate warnings and assess the evidence supporting a classification. A final Scan execution \& coverage section summarises execution across the stored scan, including all selected epochs. For each requested scanner, it reports completed transcripts against the recorded scope, or recorded scan attempts when scope information is unavailable. It also lists recorded analysis errors and available model usage from the scanners, grouped by scanner and model. Completion does not establish that usable labels were produced.

\paragraph{Dataframes.}
The returned \texttt{TransectResults} object contains pandas dataframes built from stored scanner results. These tables supply the report and can be queried, aggregated or exported for further analysis. Its \texttt{frames()} method returns the built-in tables by name. Custom tables are held separately in \texttt{layer\_frames}, while optional \texttt{turn\_tags} supports report filtering and grouping (\Cref{app:enrichment}).

\paragraph{Rows and join keys.}
Each table has a defined row unit, such as a model turn, phase, sub-agent span or a judge's record for a turn or span. Combine tables using transcript identity and the relevant turn, phase or span identifiers. Per-judge tables also distinguish the model and repetition number. When comparing separate analyses of the same transcript, retain the scan identity as well: the transcript identifier alone does not distinguish analysis configurations. A sample identifier alone is also insufficient because it can be shared across attempts. Before aggregating joined tables, check whether the join repeats records, for example when one turn has records from several judges or repetitions.

\paragraph{Table map.}
\Cref{tab:columns} summarises the returned tables, their row units and the fields used to link them. For built-in tables, combine the listed fields with \texttt{transcript\_id} within an analysis. Custom layers define their own table schemas.

\begin{table}[h]
\centering
\small
\begin{tabularx}{\linewidth}{@{}X X X@{}}
\toprule
\textbf{Table} & \textbf{One row represents} & \textbf{Linking fields within a transcript} \\
\midrule
\texttt{transcript\_info} & Transcript identity and setup & No additional field \\
\texttt{token\_timeline}, \texttt{phase\_turns} & A model turn: token records or phase attribution & \texttt{turn} \\
\texttt{flushes}, \texttt{interventions} & A recorded or inferred compaction, or human-input event & \texttt{turn}; several events can share a turn \\
\texttt{lane\_activity} & Sub-agent tool activity at a turn & \texttt{turn}, \texttt{agent\_span\_id} \\
\texttt{phases} & A behavioural phase & \texttt{phase\_index}; \texttt{turn\_start}, \texttt{turn\_end} locate its range \\
\texttt{turn\_groups} & A group of turns within a phase, with narration or fallback text & \texttt{phase\_index}, \texttt{group\_index} \\
\texttt{phase\_turn\_votes} & A judge member's record for a turn included in the digests & \texttt{turn}, \texttt{model}, \texttt{roll} \\
\texttt{subagents} & A recorded sub-agent span & \texttt{agent\_span\_id} \\
\texttt{subagent\_votes} & A judge member's record for a sub-agent span & \texttt{agent\_span\_id}, \texttt{model}, \texttt{roll} \\
\texttt{label\_definitions} & A category and its definition & \texttt{surface}, \texttt{label} \\
\texttt{turn\_tags} (optional) & Custom tags associated with a turn & \texttt{turn} \\
\texttt{layer\_frames} & Custom tables with layer-defined row units & Defined by the layer \\
\bottomrule
\end{tabularx}
\caption{Table units and linking fields. \texttt{surface} identifies the phase, sub-agent or custom-layer vocabulary. A judge member is identified by its model and repetition number (\texttt{roll}). Linking fields locate related records; they are not necessarily unique row keys. \texttt{layer\_frames} is a collection of custom tables, and \texttt{turn\_tags} is optional; both are separate from the built-in tables returned by \texttt{frames()}.}
\label{tab:columns}
\end{table}

\paragraph{Missing data and label status.}
Interpret missing values using the field's meaning and the scanner's status. An absent token count differs from a recorded zero; missing judged output may mean the scanner was not run, a call failed or a judge refused to answer. A turn can have a phase label even when its content was not directly judged. Consult \texttt{basis} and status fields where available, alongside the scan records. Some derived token measures treat missing components as zero, so check their definitions before calculating totals or averages. An absent event record does not establish that the event did not occur.

\paragraph{Schema and provenance.}
The built-in tables carry a \texttt{schema\_version} describing their table format. This field does not identify the package revision or complete analysis configuration. Reproducing an analysis also requires its input, configuration and software records (\Cref{app:reproducibility}).

\paragraph{Comparing and aggregating runs.}
Comparisons of activity or token allocation across runs require compatible category definitions and token-accounting conventions. Record differences in source coverage and judge settings, and distinguish deliberate experimental changes from differences that could confound the comparison. Phase indices locate phases within a transcript; they do not identify comparable activities across runs.

The research question and sampling design determine the statistical unit. Turns within a run and repeated judgements of those turns do not provide additional independent runs. Repeated attempts at the same task can also share task-specific influences. Depending on the question, an analysis may use run-level summaries, model dependence among observations, or both. A single transcript does not provide an empirical estimate of variation between transcripts. Specify the treatment of dependence and any multiple-comparison adjustment as part of the analysis design.

\transect{} computes descriptive summaries and judge diagnostics to support human review. Judgements of agent performance remain with the human reviewer. Inference about capability differences across agents also requires evidence that the measures capture the intended capabilities, and a separate statistical analysis. Hierarchical modelling \citep{luettgau2025hibayes} and adaptive-stopping methods \citep{pilditch2026optstop} address complementary questions about uncertainty and evaluation design.

\paragraph{Regenerating a report.}
Users can regenerate a report without rerunning the scanners or calling judge models. \texttt{load()} rebuilds the dataframes from a stored Scout scan, and \texttt{render()} generates the report. Supply the exact scan directory to select a particular analysis; when given a parent scans directory, \texttt{load()} selects the scan directory with the latest modification time. Custom layers must be supplied again through \texttt{extra\_layers} so their stored results are included (\Cref{app:enrichment}).

Rendering reads the source transcripts for excerpts and full sub-agent task prompts, and uses source tool-call counts when they are absent from the stored analysis. If these sources are unavailable, the stored analysis can still be rendered, but excerpts and any other content requiring a source read will be absent. Available excerpts are embedded in the report and can be expanded without a running Scout viewer. Optional full-transcript links require source access and a running Scout viewer, enabled through \texttt{viewer=True}. Keep the scan store and source transcripts to support later regeneration; exported analysis tables alone are not the input expected by \texttt{load()}.

\subsection{Configuring an evaluation family}
\label{app:calibration}

Family configuration defines which activities the analysis should distinguish and supplies the phase judges with relevant task context. Evaluation task descriptions, harness documentation and environment repositories provide a starting point.

\paragraph{The family Spec.}
The Spec holds a phase vocabulary, sub-agent categories and task context. Each category can include a description of the activity it covers. These definitions guide classification; they do not prescribe the order in which an agent must perform the activities. The built-in sub-agent classifier receives the category definitions but does not receive \texttt{Spec.context}. Users can also place custom analysis content in \texttt{Spec.extra}, which their custom layers must interpret (\Cref{app:enrichment}).

\paragraph{Expert elicitation.}
Structured expert elicitation \citep{hanea2017idea, ohagan2006uncertain} can help users address gaps in evaluation materials and review proposed category definitions. For example, experts can identify outcomes worth investigating and clarify the boundary between planning an experiment and carrying it out. Users can incorporate the resulting definitions and context into the Spec. Users can prepare questions and draft rubrics for review and revision with relevant experts.

Users should record which sources and expert contributions informed the rubric, alongside its validation status and supporting evidence (\Cref{app:trust}). The package retains the vocabulary but does not maintain a dedicated record of expert contributions. Expert review can inform a rubric without establishing how accurately its classifications describe agent behaviour.

\paragraph{Analysis settings.}
Retain the Spec and invocation settings together to distinguish changes in behavioural definitions from changes in their application. Assessing a revised rubric or judging configuration requires a new analysis (\Cref{app:reproducibility}).

\paragraph{Developing the vocabulary.}
Development transcripts can help users identify activities to include in a preliminary vocabulary. Review these examples alongside evaluation materials and expert input, recording which sources informed each category and its definition. No fixed number of transcripts guarantees coverage: behaviours absent from the examples may still matter in later runs.

Keep the transcripts used to develop or revise the vocabulary separate from held-out data used to assess its performance or generality. A vocabulary can be explored on the examples that informed it, but those results do not provide an independent assessment. If findings from held-out transcripts inform a revision, treat those transcripts as development evidence and use data that did not inform the revision for an independent assessment. These are research-design practices; the package does not enforce the separation.

The built-in phase and sub-agent scanners include \texttt{none\_of\_the\_above} even when it is absent from the Spec. Inspect these labels for possible gaps in the vocabulary. The fallback does not guarantee that every omitted behaviour will be detected. Custom scanners using \texttt{cohort\_llm\_scanner} classify against the supplied label list; the helper does not add a fallback category.

\paragraph{Case-study vocabularies.}
The CRUX case in \Cref{sec:cases} uses two vocabularies for custom judged analyses of orchestrator activity. The research-activity rubric focuses on general activities common to any AI R\&D task. \textit{Hypothesising} proposes a new mechanism, approach or explanation \cite{si2025can,simon1989scientist,peirce1934collected}, while \textit{Literature grounding} uses prior work to inform that direction \cite{skarlinski2024language}. \textit{Experiment design} specifies what to vary and measure \cite{fisher1966design}; \textit{Analysis design} chooses the statistical instrument for interpreting the evidence \cite{gelman2016statistical,silberzahn2018many,steegen2016increasing}. \textit{Falsification} covers designing or executing a test intended to disconfirm the agent's own hypothesis, with a failure condition set before the relevant results are seen \cite{chamberlin1890method,mayo1996error,platt1964strong}. Once work is specified, \textit{Research engineering} implements, runs or monitors it \cite{chan2025mle,starace2025paperbench,wijk2024rebench}, and \textit{Debugging} addresses broken code \cite{tian2024debugbench}. \textit{Evidence interpretation} connects an observed result to its scientific implications \cite{chinn1993role,kulkarni1988processes}, while \textit{Reflecting} reassesses progress or strategy beyond a single result \cite{dunbar1994scientists}. \textit{Writing up} brings findings into the manuscript through drafting, revision and review, including figures, bibliography checks and compilation. A residual \textit{Other} label covers generic operations, insufficient evidence, or work outside the vocabulary.

The paper-specific vocabulary focuses specifically on the TabPFN task's concrete research questions and deliverables. \textit{PFN problem framing} addresses harmful versus benign shift under the no-label deployment constraint; \textit{PFN literature grounding} and \textit{PFN solution ideation} cover positioning against prior detectors and developing ideas that exploit PFN affordances. Method development spans implementation (\textit{PFN detector construction}) and theoretical guarantees (\textit{PFN theory}); evaluation requires building the apparatus (\textit{Benchmark construction}) and running detectors and baselines (\textit{PFN model evaluation}). \textit{Running ablations} isolates the source of a signal, while \textit{Falsification tests} challenge the detector's core claim against a named failure condition. Turning results into a deliverable brings together \textit{Manuscript write-up}, \textit{Self-review} against conference criteria, and \textit{Code reproducibility} through runnable analyses. The vocabulary also tracks \textit{Non-research-relevant engineering}, such as provisioning, permissions and file management, making operational work visible alongside these research activities. The case-study classification also includes a residual \textit{Other} label for outputs outside these named categories.

\subsection{Judging and auditing}
\label{app:trust}

The judging regime determines which consistency checks are available. A \emph{solo} configuration uses one judge model for an initial judging pass over the classification inputs. A \emph{$k$-roll} configuration repeats the same model, providing evidence about within-model stability. A \emph{cohort} configuration uses several models, providing evidence about between-model agreement. The package keeps repeated calls to one model separate from multi-model voting: a cohort cannot also set \texttt{k\_rolls} above one. Each member is identified by its model and repetition number.

Agreement can reflect shared errors, so neither repeated-call stability nor cross-model agreement establishes correctness. A solo judgement can carry stated confidence but supplies no repeated-judgement estimate of consistency. Verifier reviews and narrative generation can add further model calls. Cohorts can include models from different families. Model diversity does not guarantee independent errors or correct judgements.

\paragraph{Combining judgements.}
Voting selects the most frequent usable label for each unit. Phase classification combines votes at the judged-turn level before constructing phases; sub-agent classification combines votes for each classified span. If label counts tie, the tied label with the higher mean confidence wins, treating missing confidence as zero for this tie-break. A remaining tie is resolved by member order.

Vote agreement is the fraction of usable votes supporting the selected label. Refusals, failed calls, missing judgements and filled phase labels do not contribute votes, so agreement must be read alongside the number of usable voters and the recorded failures. One usable voter supplies no evidence of agreement between members. The available member records retain individual judgements and recorded failure states for inspection. Filled turns were included in the digests but inherit an earlier label when the returned phase segments leave them uncovered. Attributed turns lie outside the digests and are mapped onto the phase timeline.

The vote's confidence is the mean of the available stated confidences among members supporting the selected label. It is distinct from vote agreement and is not a calibrated probability that the label is correct. Vote agreement is not chance-corrected. Where verification is enabled, a later review may change the final label. Distinguish the original voting results from the verifier's subsequent assessment.

\paragraph{Selecting judgements for verifier review.}
The verifier adds a model review of selected classifications. It runs by default in solo and $k$-roll configurations; the package rejects verification with a multi-model cohort. Analysis settings can disable verification or select a verifier model; otherwise, the first configured judge model is used.

A phase is selected for review when the lowest confidence among its judged or filled digest turns is below 0.6, or the lowest available turn agreement is below 0.6. Review also covers phases containing at most two such digest turns between neighbouring phases that share a different label. For sub-agent classification, the triggers are stated confidence below 0.6 or vote agreement below 0.6 where available. These thresholds select candidates for review; they do not identify proven errors.

Both scanners also select additional judgements for spot checks. By default, the phase scanner targets the larger of three phases or 5\% of all phases, rounded to a whole number, drawing from the remaining phases in label-wise rounds with a fixed seed. The target is capped by the number available. The sub-agent scanner uses a deterministic per-item draw with a default 5\% selection probability, without a minimum count. \texttt{verify\_sample} changes the spot-check setting; an explicit phase setting uses the requested fraction without the default minimum of three.

\paragraph{Verifier decisions and records.}
The verifier sees the earlier classification. For phases, it also receives selected turn digests and neighbouring phase labels; for sub-agent classifications, it reviews the classification input alongside the earlier answer and explanation. It proposes a label from the same vocabulary. A different label is applied only when the verifier's stated confidence is at least 0.6. Phase changes are applied to the affected turn judgements, after which phases are reconstructed.

Available review records identify why a judgement was selected, the original and proposed labels, the verifier's confidence and whether the change was applied. A review without a usable verdict leaves the earlier label unchanged. Sub-agent review records distinguish refusals and errors; phase verification counts phases without a usable verdict. After phase reconstruction, merged phases preserve all recorded reviews of their constituent original phases in \texttt{verifier\_reviews}. A verifier that sees the original answer is not a blind independent assessment, even when it uses a different model. Agreement with the first judgement does not establish correctness, and an applied label change does not by itself establish improvement. A label-change rate measures model revisions, not independently established errors. Targeted selection and the different spot-check schemes also limit generalisation from reviewed cases to the run.

\paragraph{Reading the audit.}
For each included judged analysis, the reliability audit shows the judging regime, available confidence and agreement measures, verifier information, and breakdowns by label or judge where the required records are present. These views help reviewers distinguish disagreement about a label from missing judgements or changes introduced by verification.

Warnings can direct readers to the audit for more detailed diagnostics. For example, low agreement alongside poor member coverage warrants checking which judges returned usable labels before changing the rubric. A warning identifies a condition to investigate; its suggested explanations or follow-up actions do not establish the cause. Custom-layer audit participation is optional; \Cref{app:enrichment} describes the supported interface and recommended checks.

\paragraph{Agreement and member coverage.}
The package reports several agreement measures with different meanings. Per-unit vote agreement is the share of votes supporting the selected label, as defined above. Pooled pairwise agreement instead counts agreeing pairs of recorded member labels, divided by all within-unit pairs, pooling units with at least two non-missing labels. Units with more recorded labels contribute more pairs.

The package also computes nominal Krippendorff's $\alpha$ and Gwet's AC1, which compare observed agreement with different models of chance agreement. The AC1 calculation uses pooled pairwise agreement and category proportions, counting only categories observed in the eligible ratings. These are consistency measures, not classification accuracy. They use units with at least two recorded labels; the accompanying \texttt{n} counts the ratings in those units, not independent runs. The current implementation returns an unavailable coefficient when only one category is observed among eligible units, rather than treating that case as perfect chance-corrected agreement.

Member coverage reports, for each model and repetition, the fraction of its recorded unit rows that contain a judgement, alongside counts of recorded failures, missing responses and filled phase labels. Phase coverage counts directly judged rows; sub-agent coverage counts rows with an \texttt{ok} status. Coverage is conditional on the records present and cannot detect a member or unit omitted entirely from them. Read coverage alongside agreement: high agreement can coexist with low coverage.

\paragraph{Label-level summaries.}
The audit breaks down label counts, available confidence and agreement values, and the recorded source of each label by category. Phase summaries use directly judged turn rows; sub-agent summaries use spans with a recorded label. When the declared vocabulary is available, unused categories remain visible. An unused category means no included unit received that label; it does not establish that the corresponding behaviour was absent.

The minority-vote statistic asks how often a member's use of a particular label differs from the selected label for the same unit. Its denominator is the recorded member votes for that label that can be matched to a decided unit; units whose final label came from the verifier are excluded. The provenance breakdown counts which recorded label source produced each decided label, such as a single judge, a vote or a verifier. These counts describe how decisions were made, not how many were correct.

\paragraph{Verifier-change summaries.}
For current built-in scanner outputs, the overall verifier-change rate is the fraction of completed reviews with usable verdicts that resulted in an applied label change. Per-label rates group these reviews by their original label; the spot-check rate uses completed reviews selected by the \texttt{random\_sample} trigger. Phase summaries count reviews of the original phases, preserving them through later merging. Reviews without usable verdicts are counted separately and excluded from these rate denominators. These are summaries of the stored review records, not necessarily of every attempted review.

Read these rates with the selection and record-coverage limits above. A low change rate can reflect agreement with the initial labels or proposed changes below the confidence threshold for application; it does not establish a low error rate.

\paragraph{Intervals on diagnostic summaries.}
For counted proportions, including member coverage and verifier-change rates, the package uses nominal 95\% Wilson intervals and returns no rate or interval when the denominator is zero. For audit summaries of mean confidence or agreement, it drops missing values and uses the normal approximation $\bar{x} \pm 1.96s/\sqrt{n}$, where $s$ is the sample standard deviation and $n$ is the number of retained values. These audit mean intervals are omitted below eight values.

These calculations do not account for dependence between turns, shared judge errors or the verifier's selection process. Their nominal level therefore does not guarantee 95\% coverage for a claim about a run or evaluation family. An interval around mean stated confidence describes those confidence values; it does not turn them into calibrated probabilities of correctness.

\paragraph{Scanner-level confidence summaries.}
Scanner outputs also contain confidence summaries calculated over different sets of values. A vote's \texttt{confidence\_pm} applies the normal-approximation half-width $1.96s/\sqrt{n}$ to the available stated confidences of members supporting its selected label. The helper returns zero when fewer than two values are available. This differs from the audit mean interval, which is omitted below eight values.

For phases, \texttt{confidence} averages the confidence values of the phase's judged or filled digest turns, treating a missing value as zero. \texttt{confidence\_spread} applies the same half-width helper to those turn values. This is a half-width for their mean, not their standard deviation or a measure of disagreement between judges. After an applied verifier change, affected turns receive the verifier's confidence, and the resulting verifier-marked phase has its spread set to zero. A zero spread can therefore reflect insufficient values or assignment after verifier review, as well as identical confidence values; it is not evidence of certainty. 

\paragraph{Warning thresholds.}
The current report uses fixed rules to flag diagnostics for inspection:
\begin{itemize}
    \item Mean $k$-roll vote agreement below 0.80 or above 0.95 produces an amber warning.
    \item Cohort Krippendorff's $\alpha$ below 0.66 produces a red warning; values from 0.66 up to, but not including, 0.80 produce an amber warning.
    \item Overall or per-label verifier-change rates of at least 0.20 produce an amber warning.
    \item Any applied change in the recorded spot-check subset produces a red warning.
    \item Where a judging regime is recorded, missing classifications produce an amber warning when their share is above zero but below 25\%, and a red warning at 25\% or more. For phases, this share counts \texttt{refusal}, \texttt{no\_answer} and \texttt{missing\_turn} rows among all non-attributed phase-turn rows; filled turns enter the denominator only. For sub-agent classification, it counts rows without a label among all recorded classification rows in the \texttt{sub-agents} table, including structurally recorded spans with no classification result.
\end{itemize}
These rules apply where the required metrics are available. The report also groups stated confidence into low (below 0.66), medium (0.66 to below 0.80) and high (at least 0.80) bands. These are display conventions, not calibrated levels of correctness.

Threshold crossings direct attention to the audit and source records. High self-agreement does not by itself diagnose label-name bias, and a spot-check label change does not prove that the original judgement was wrong. Conversely, the absence of a warning does not establish that the classifications are reliable or valid.

\paragraph{Evidence for interpreting classifications.}
Software tests can establish whether a scanner extracts or processes information as specified on the tested inputs. Repeated judgements and cross-model comparisons assess consistency. Neither establishes that a behavioural category captures the intended construct or that its labels are correctly applied to new transcripts.

External assessment compares classifications with an appropriate reference, such as independently collected expert annotations. Its interpretation depends on the reference's quality, the assessment procedure and the transcripts represented. Expert review of category definitions is useful for developing a rubric, but is distinct from assessing how judges apply it. A comparison with another model provides additional model evidence; it does not become ground truth merely because that model is different or more capable.

\paragraph{Testing sensitivity to category names.}
The \texttt{scramble\_spec} helper creates a copy of a Spec with neutral identifiers substituted for its phase and sub-agent labels. It returns a mapping for each classification type that \texttt{descramble} uses to translate output labels back to the original names. Descriptions, task context and operational-category flags are preserved. The helper does not rewrite label names mentioned inside those descriptions or the context, and it does not rename custom vocabularies in \texttt{Spec.extra}. Check which category names and definitions the judge actually receives, including reserved categories, before interpreting a comparison.

Users run the original and renamed configurations on the same inputs and judging settings, then compare outputs for the same turns or sub-agent spans after translating the labels back. Repeated unchanged runs help distinguish sensitivity to names from ordinary judging variation; cached responses cannot provide fresh evidence of that variation. Differences can motivate closer inspection of label wording or rubric definitions, but do not by themselves establish undue reliance on category names. Stability under renaming does not establish classification accuracy. Renamed runs are diagnostic artefacts and should be clearly identified as such; the package does not automatically mark their reports. 

\paragraph{Ablations and other perturbations.}
An ablation removes a specified part of an analysis to test what depends on it. For example, removing an expert-supplied contextual instruction while retaining the transcript, category definitions and other analysis settings asks how much the classifications depend on that instruction. Before running the comparison, state what change is expected and which output will be compared. Keep other relevant conditions matched and compare the difference with variation under an unchanged configuration.

Check that the intended change reached the configuration or prompt actually used by the analysis. Compare outputs for matching source units, such as turns or sub-agent spans. Report failures and missing coverage in each condition alongside differences between conditions for matched units with usable outputs in both.

Removing a category also changes the available answers, so a resulting label change cannot be attributed solely to the removed category's usefulness. Removing a report layer addresses a different question: its contribution to human review requires an appropriate reviewer study. Perturbations that disrupt transcript information or ordering likewise need an explicit prediction; preserving or losing a signal is not automatically the desirable outcome.

These are user-designed comparisons, supported by configurable inputs and stored outputs. The package does not provide a general ablation runner or automatically validate custom scanners through perturbation tests. Any claim that one configuration is more accurate needs an appropriate external assessment.

\paragraph{Scope of the case study.}
The case study illustrates how the combined report supports investigation of an agent's research process. It does not provide a controlled estimate of reviewer time saved, findings missed or improvements in review decisions. Nor does it isolate the contribution of behavioural classification from that of making a long transcript navigable. Claims about classification accuracy or reviewer benefit require evidence tied to the relevant rubric, package version and evaluation setting.

\paragraph{Documenting validation evidence.}
Users should document what their categories are intended to represent, what assessment has been performed, the reference and data used, and the conditions to which the evidence applies. Distinguish expert review of category definitions from assessment of their application, and state when evidence is absent, preliminary or limited in scope. The package does not maintain a dedicated validation-status record; users must retain this documentation separately. Recording a status does not establish validity.

\subsection{Extending the analysis}
\label{app:enrichment}

Custom scanner outputs can be converted into dataframes for further analysis or use in timeline layers and other report elements. Users can also model or analyse the dataframes generated by \transect{} and incorporate the resulting figures or other outputs as custom report elements. Users pass \texttt{Layer} objects through \texttt{extra\_layers}. A layer can specify a scanner and a function that builds its dataframe, or accept an existing dataframe. It can also define report elements and phase-card tags. A custom report element therefore need not introduce another scanner. Custom tables are available through \texttt{results.layer\_frames}, keyed by layer name. The returned dataframes also support separate dashboards, figures or summaries.

\paragraph{Custom scanners.}
Custom scanners can extract structural information or ask judge models to classify behaviour. For example, a structural scanner might identify recorded test failures, while a judged scanner might classify research activities using a family-specific rubric. Each scanner is registered with Scout's \texttt{@scanner} decorator and supplied through a layer. Its results are saved in the Scout scan store and can be converted into a custom dataframe.

For classifications of individual items, \texttt{cohort\_llm\_scanner} provides shared judging support (\Cref{app:trust}), including voting, optional verification and judge metadata. Tasks such as phase segmentation need additional logic to carry context between items; the built-in \texttt{decision\_phases} scanner provides an implementation reference. Reusing the helper does not establish the validity of a custom rubric.

For custom classifications using \texttt{cohort\_llm\_scanner}, a loader can group several units into a window. With \texttt{batch=True}, the scanner judges that window in one initial call per member and retains separate results for each unit. Voting and any verifier review operate per unit. Results identify units with missing answers. For example, \texttt{reasoning\_turns(batch=N)} with $N>1$ groups up to $N$ eligible main-agent turns. This loader's digests contain available recorded reasoning, message text, tool names and delegation information. Turns containing only readable reasoning are eligible. The loader excludes recognised OpenClaw failure placeholders and turns with no text, readable reasoning or delegation information. Custom loaders can define different classification units and eligibility rules.

We recommend provenance and reliability review for every custom scanner. Record its inputs and method, the available evidence and any gaps. For structural scanners, document source coverage and extraction settings, distinguishing extracted, inferred and unavailable values; repeatability alone does not establish complete or correct extraction. The package offers an opt-in report audit for judged layers whose dataframes satisfy its audit contract. Other outputs require checks appropriate to their data and intended use; the package does not enforce universal audit coverage.

\paragraph{Custom report elements.}
A layer's \texttt{section} defines how its results appear in the report. Built-in report blocks display numeric turn-by-turn charts, categorical bands, spans, event markers, sortable tables and explanatory text. For further customisation, \texttt{Component} calls a user-supplied function that returns an Inspect Viz component. These interfaces let users present figures or other outputs from their own modelling or analysis of \transect{}'s dataframes. Turn-based displays use the shared timeline; a table or other summary need not be organised by turn.

Blocks with dataframe inputs can read the layer's dataframe or an explicitly supplied dataframe. If the dataframe includes \texttt{transcript\_id}, rendering selects rows for the current transcript; otherwise it reuses the supplied data for each transcript. A custom \texttt{Component} function receives the current transcript's context and must select any additional data it uses appropriately. Users should preserve transcript identity when an analysis is run-specific, so its outputs are not inadvertently repeated in other reports.

\paragraph{Custom report controls.}
A layer's \texttt{tags} setting maps categorical columns in a per-turn dataframe to named tag groups. These tags appear on phase cards, provide filters for finding relevant episodes, and add grouping choices to the token-use display. For example, a custom research-activity classification can let reviewers filter cards by activity and inspect token use under the same categories. Numeric analysis outputs can instead be shown in charts or tables.

Users set \texttt{section\_order} when generating or regenerating a report. It places the listed sections in the requested order, then appends any remaining sections in their default order. Each transcript's reliability audit follows its ordered sections, and the run-wide Scan execution \& coverage section closes the report. This controls report layout without changing the underlying classifications. The optional \texttt{sensitivity} setting adds a user-supplied protective marking to the report title and fixed banners at the top and bottom. No marking is added by default.

\paragraph{Reading signals together.}
The shared timeline lets reviewers examine several aspects of the same episode. For example, they can inspect a context-compaction marker alongside a change in the activity label and a rise in token use, then return to the transcript to investigate what happened. The source records and scanner audit help reviewers assess how each signal was produced and what its available diagnostics support (\Cref{app:trust}). A missing signal may reflect incomplete logging, a scanner that did not run, or a classification failure; it does not by itself show that the behaviour was absent.

Signals appearing near one another on the timeline do not establish a causal relationship. Nor do several outputs necessarily provide independent evidence: they may share transcript excerpts, category definitions, judge models or derived values. For example, token use grouped by phase depends on the phase labels and cannot independently verify those labels.

Related signals can still help reviewers identify patterns, inconsistencies and questions for follow-up. When drawing a conclusion from several signals, document their shared inputs and assumptions and assess what evidence each adds. The report does not automatically measure dependence or establish the validity of the conclusion.

\subsection{Reproducing and scaling analyses}
\label{app:reproducibility}

\paragraph{Records to retain.}
Retain the source transcripts, selected sample and epochs, family Spec, judging settings, custom-layer definitions and exact scan directory. Record the package revision and relevant dependency and provider-model versions where available, including local code changes. Together, these records distinguish changes in the evaluation evidence from changes in how it was processed.

Stored scanner results and dataframes carry useful provenance, including recorded judge identities, label definitions and decision sources. They do not replace the complete analysis configuration or a record of the software environment.

\paragraph{Stored results, cached responses and fresh judgements.}
Reloading a scan with \texttt{load()} rebuilds its dataframes from stored scanner results without calling judge models. A new \texttt{transect()} invocation performs another scan, but judge calls can reuse Inspect's response cache when their cache keys match. Repeated calls within a $k$-roll configuration use separate roll-specific cache scopes, so a later run can reuse the response from each corresponding roll. A new scan directory alone does not ensure fresh judge responses.

For each repeat intended to assess judging variation, set \texttt{INSPECT\_CACHE\_DIR} to a new, empty directory before starting its analysis process. Verify that relevant calls did not replay cached responses. Caching remains active within each process. This setting controls Inspect's model-response cache; custom scanners may have other reuse mechanisms.

Structural extraction makes no model calls, but its outputs can change with the source data, adapter or scanner implementation. Fresh judged outputs may vary even with unchanged settings; identical outputs are also possible and do not establish that a model or pipeline is deterministic.

Each \texttt{transect()} invocation on OpenClaw JSONL input imports the supplied files into a fresh transcript snapshot database under the scans directory. Duplicate transcript identifiers across the supplied files raise an error. Loading a stored scan does not repeat the import.

\paragraph{Comparing analysis revisions.}
Before comparing outputs, check that they refer to the same source transcripts and align the relevant sample, epoch, turn or span identifiers. Record changes to the vocabulary, input processing and judging settings. A change in category definitions or turn numbering can make a direct label comparison misleading.

When comparing configurations whose judged outputs can vary, estimate unchanged-configuration variation using fresh judgements under the relevant conditions. This provides context for a changed configuration; it does not supply a universal tolerance band or establish that a difference is an improvement. Comparisons intended to establish greater classification accuracy need an appropriate external assessment.

\paragraph{Documenting configuration choices.}
Document why a family uses particular categories, thresholds and judge settings, including their sources, assumptions and changes between analyses. Keep expert-supplied choices distinguishable from defaults and generated suggestions. The Spec and interface checks constrain the form of the configuration; they do not establish the scientific justification for its contents. Keep this documentation alongside the Spec and stored analysis, including the supporting evidence and its limits.

\paragraph{Assistance for users.}
The package ships skills for coding assistants that guide common tasks. \texttt{using-transect} covers Spec authoring, running an analysis, reading reports and analysing dataframes. \texttt{transect-diagnostics} guides investigation of disagreements, missing outputs and judging settings, including the label-renaming diagnostic. \texttt{add-a-layer} explains how to add scanners, report elements, tags and audit entries. \texttt{custom-ui} guides separate presentations of the results, including how to retain provenance, judging diagnostics and missing-data information. These skills provide instructions and examples; users remain responsible for selecting behavioural categories, reviewing proposed configurations and definitions, recording their sources and assumptions, and interpreting results.

Running \texttt{python -m transect.skills install} from the project directory copies the packaged skills into \texttt{.claude/skills} for Claude Code. Development-repository hooks for regression checks and other working practices are separate from this installer. They should not be assumed to run in a user's package installation or another coding-assistant environment.

\paragraph{Cost and runtime.}
Judged-analysis cost depends on input text, the number of classification units, judging regime, verifier coverage, narration and response-cache reuse. Longer transcripts may require more phase-classification chunks; additional sub-agent spans or custom items may require more classifications. Repeated calls and cohort members add judging work, while cached responses can avoid provider requests. API charges depend on the chosen models' input, output and cache prices. Tokens used for analysis are separate from the evaluated agent's token totals.

Structural analysis requires computation, storage and report generation without calling judge models. Concurrency, batching and provider limits also affect runtime, which cannot be inferred directly from token counts or API charges. Custom-scanner batching groups initial classifications into shared calls (\Cref{app:enrichment}); verifier reviews can add calls. No speed or cost advantage is established here.

\paragraph{Scaling oversight.}
Users can add scanners, relevant context or role-specific rubrics, or choose alternative judge models, to address further evaluation questions through the shared tables and report. These additions can also increase cost, disagreement and shared errors; more context or more judges need not improve classification quality.

Provenance and reliability audits help users inspect outputs, identify missing evidence and select findings for human review. Users can then assess whether further analysis is worth its cost and what confidence the evidence warrants. Demonstrating improved accuracy, reviewer performance or cost-effectiveness requires an appropriate comparison.

%% file: references.bib
@article{kapoor2026openworld,
  title        = {Open-World Evaluations for Measuring Frontier {AI} Capabilities},
  author       = {Kapoor, Sayash and Kirgis, Peter and Schwartz, Andrew and Rabanser, Stephan and Allaire, J. J. and Bommasani, Rishi and Coppock, Harry and Dubois, Magda and Hadfield, Gillian K. and Hall, Andrew B. and Hooker, Sara and Lazar, Seth and others},
  year         = {2026},
  journal      = {arXiv preprint arXiv:2605.20520}
}

@article{dubois2026sevensteps,
  title        = {Seven Simple Steps for Log Analysis in {AI} Systems},
  author       = {Dubois, Magda and Zorer, Ekin and Hamin, Maia and Skinner, Joe and Souly, Alexandra and Wynne, Jerome and Coppock, Harry and Sato, Lucas and Kapoor, Sayash and others},
  year         = {2026},
  journal      = {arXiv preprint arXiv:2604.09563},
}

@article{folkerts2026cyber,
  title        = {Measuring {AI} Agents' Progress on Multi-Step Cyber Attack Scenarios},
  author       = {Folkerts, Linus and Payne, Will and Inman, Simon and Giavridis, Philippos and Skinner, Joe and Deverett, Sam and Aung, James and others},
  year         = {2026},
  journal      = {arXiv preprint arXiv:2603.11214},
}

@article{luettgau2025hibayes,
  title        = {{HiBayES}: A Hierarchical {Bayesian} Modeling Framework for {AI} Evaluation Statistics},
  author       = {Luettgau, Lennart and Coppock, Harry and Dubois, Magda and Summerfield, Christopher and Ududec, Cozmin},
  year         = {2025},
  journal      = {arXiv preprint arXiv:2505.05602}
}

@techreport{nist2026toolbox,
  title       = {Expanding the {AI} Evaluation Toolbox with Statistical Models},
  author      = {Keller, A. and Kwegyir-Aggrey, K. and Steed, R. and Rao, A. and Sharp, J. and Bergman, A.},
  year        = {2026},
  institution = {National Institute of Standards and Technology (NIST), Gaithersburg, MD},
  number      = {NIST AI 800-3},
  type        = {NIST Trustworthy and Responsible AI},
  doi         = {10.6028/NIST.AI.800-3}
}

@misc{transluce_docent,
  title        = {Docent: A Behaviour-Analysis Platform for {AI} Agents},
  author       = {{Transluce}},
  year         = {2026},
  howpublished = {\url{https://docs.transluce.org/introduction}},
  note         = {Accessed 2026-07-01}
}

@article{wijk2024rebench,
  title   = {{RE-Bench}: Evaluating Frontier {AI} {R\&D} Capabilities of Language Model Agents against Human Experts},
  author  = {Wijk, Hjalmar and Lin, Tao and Becker, Joel and Jawhar, Sami and Parikh, Neev and Broadley, Thomas and Chan, Lawrence and Chen, Michael and Clymer, Josh and Dhyani, Jai and Ericheva, Elena and Garcia, Katharyn and Goodrich, Brian and Jurkovic, Nikola and Karnofsky, Holden and Kinniment, Megan and Lajko, Aron and Nix, Seraphina and Sato, Lucas and Saunders, William and Taran, Maksym and West, Ben and Barnes, Elizabeth},
  year    = {2024},
  journal = {arXiv preprint arXiv:2411.15114},
  note    = {METR. Also published at ICML 2025, PMLR 267:66772--66832.}
}

@article{starace2025paperbench,
  title   = {{PaperBench}: Evaluating {AI}'s Ability to Replicate {AI} Research},
  author  = {Starace, Giulio and Jaffe, Oliver and Sherburn, Dane and Aung, James and Chan, Jun Shern and Maksin, Leon and Dias, Rachel and Mays, Evan and Kinsella, Benjamin and Thompson, Wyatt and Heidecke, Johannes and Glaese, Amelia and Patwardhan, Tejal},
  year    = {2025},
  journal = {arXiv preprint arXiv:2504.01848},
  note    = {OpenAI.}
}

@misc{aisi2026compute,
  title        = {More Compute, More Capability: Why {AI} Agent Evaluations Need to Account for Test-Time Compute},
  author       = {{UK AI Security Institute}},
  year         = {2026},
  howpublished = {AISI blog (Science of Evaluations team)},
  url          = {https://www.aisi.gov.uk/blog/more-compute-more-capability-why-ai-agent-evals-need-to-account-for-test-time-compute},
}

@misc{mcfadyen2026inferencecomputeshapesfrontier,
      title={How Inference Compute Shapes Frontier {LLM} Evaluation}, 
      author={Jessica McFadyen and Ole Jorgensen and Harry Coppock and Kevin Wei and Cozmin Ududec},
      year={2026},
      eprint={2606.17930},
      archivePrefix={arXiv},
      primaryClass={cs.AI},
      url={https://arxiv.org/abs/2606.17930}, 
}

@misc{aisi2026cheating,
  title        = {Cheating Behaviour in Frontier Model Evaluations},
  author       = {{UK AI Security Institute}},
  year         = {2026},
  howpublished = {AISI blog},
  url          = {https://www.aisi.gov.uk/blog/cheating-behaviour-in-frontier-model-evaluations},
}

@inproceedings{reizinger2026skills,
  title        = {Skills, Benchmarks, and Verification Are What {AI}-Assisted Research Needs},
  author       = {Reizinger, Patrik and Brendel, Wieland},
  year         = {2026},
  booktitle    = {International Conference on Learning Representations (ICLR)},
}

@article{liu2026lasthuman,
  title        = {The Last Human-Written Paper: Agent-Native Research Artifacts},
  author       = {Liu, Jiachen and Pei, Jiaxin and Huang, Jintao and Si, Chenglei and Qu, Ao and Tang, Xiangru and others},
  year         = {2026},
  journal      = {arXiv preprint arXiv:2604.24658},
}

@article{yehudai2025survey,
  title        = {Survey on Evaluation of {LLM}-based Agents},
  author       = {Yehudai, Asaf and Eden, Lilach and Li, Alan and Uziel, Guy and Zhao, Yilun and Bar-Haim, Roy and others},
  year         = {2025},
  journal      = {arXiv preprint arXiv:2503.16416}
}

@article{mohammadi2025survey,
  title        = {Evaluation and Benchmarking of {LLM} Agents: A Survey},
  author       = {Mohammadi, Mahmoud and Li, Yipeng and Lo, Jane and Yip, Wendy},
  year         = {2025},
  journal      = {arXiv preprint arXiv:2507.21504}
}

@article{hanea2017idea,
  title   = {Investigate Discuss Estimate Aggregate for Structured Expert Judgement},
  author  = {Hanea, A. M. and McBride, M. F. and Burgman, M. A. and Wintle, B. C. and Fidler, F. and Flander, L. and Twardy, C. R. and Manning, B. and Mascaro, S.},
  journal = {International Journal of Forecasting},
  volume  = {33},
  number  = {1},
  pages   = {267--279},
  year    = {2017}
}

@book{ohagan2006uncertain,
  title     = {Uncertain Judgements: Eliciting Experts' Probabilities},
  author    = {O'Hagan, Anthony and Buck, Caitlin E. and Daneshkhah, Alireza and Eiser, J. Richard and Garthwaite, Paul H. and Jenkinson, David J. and Oakley, Jeremy E. and Rakow, Tim},
  publisher = {John Wiley \& Sons},
  series    = {Statistics in Practice},
  year      = {2006},
  isbn      = {978-0-470-02999-2},
  doi       = {10.1002/0470033312}
}

@article{norman2026reliability,
  title   = {Reliability without Validity: A Systematic, Large-Scale Evaluation of {LLM}-as-a-Judge Models Across Agreement, Consistency, and Bias},
  author  = {Norman, Justin D. and Rivera, Michael U. and Hughes, D. Alex},
  year    = {2026},
  journal = {arXiv preprint arXiv:2606.19544},
  note    = {Submitted 17 Jun 2026.}
}

@misc{inspectai,
  title        = {Inspect {AI}: A Framework for Large Language Model Evaluations},
  author       = {{UK AI Security Institute}},
  howpublished = {\url{https://inspect.aisi.org.uk/}},
  note         = {Accessed 2026},
  year         = {2026}
}

@misc{inspectscout,
  title        = {{Inspect Scout}: A Transcript-Analysis Library},
  author       = {{Meridian Labs}},
  howpublished = {\url{https://meridianlabs-ai.github.io/inspect_scout/}},
  note         = {Accessed 2026},
  year         = {2026}
}

@misc{pilditch2026optstop,
      title={Knowing When to Stop: Bayesian Optimal Stopping for {LLM} Evaluations}, 
      author={Toby D. Pilditch},
      year={2026},
      eprint={2608.14425},
      archivePrefix={arXiv},
      primaryClass={cs.AI},
      url={https://arxiv.org/abs/2608.14425}, 
}

@misc{zhang2026stop,
  title         = {Stop Comparing {LLM} Agents Without Disclosing the Harness},
  author        = {Zhang, Yunbei and Wang, Janet and Ge, Yingqiang and Xu, Weijie and Hamm, Jihun and Reddy, Chandan K.},
  year          = {2026},
  eprint        = {2605.23950},
  archivePrefix = {arXiv},
  primaryClass  = {cs.AI},
  url           = {https://arxiv.org/abs/2605.23950}
}

@misc{yao2026harnessbench,
  title         = {{Harness-Bench}: Measuring Harness Effects across Models in Realistic Agent Workflows},
  author        = {Yao, Yilun and Tan, Xinyu and Liu, Chao-Hsuan and Li, Yaoming and Wang, Zhengyang and Yu, Wenhan and Tan, Zhewen and Tian, Yuxuan and Zhao, Guangxiang and Sun, Lin and Zhang, Xiangzheng and Yang, Tong},
  year          = {2026},
  eprint        = {2605.27922},
  archivePrefix = {arXiv},
  primaryClass  = {cs.AI},
  url           = {https://arxiv.org/abs/2605.27922}
}

@article{zhou2026general,
  title={General scales unlock {AI} evaluation with explanatory and predictive power},
  author={Zhou, Lexin and Pacchiardi, Lorenzo and Mart{\'\i}nez-Plumed, Fernando and Collins, Katherine M and Moros-Daval, Yael and Zhang, Seraphina and Zhao, Qinlin and Huang, Yitian and Sun, Luning and Prunty, Jonathan E and others},
  journal={Nature},
  volume={652},
  number={8108},
  pages={58--67},
  year={2026},
  publisher={Nature Publishing Group UK London}
}

@article{cronbach1955construct,
  title={Construct validity in psychological tests.},
  author={Cronbach, Lee J. and Meehl, Paul E.},
  journal={Psychological Bulletin},
  volume={52},
  number={4},
  pages={281--302},
  year={1955},
  publisher={American Psychological Association},
  doi={10.1037/h0040957}
}

@article{bowkis2026automated,
  title={Automated alignment is harder than you think},
  author={Bowkis, Aleksandr and Buhl, Marie Davidsen and Pfau, Jacob and Irving, Geoffrey},
  journal={arXiv preprint arXiv:2605.06390},
  year={2026}
}

@article{kirgis2026can,
  title={Can {AI} agents conduct open-ended {AI} research? {Early} evidence from two case studies},
  author={Kirgis, Peter and Kapoor, Sayash and Schwartz, Andrew and Rabanser, Stephan and Africa, David and Voudouris, Konstantinos and Nguyen, Viet and Pilditch, Toby and Dubois, Magda and Coppock, Harry and others},
  journal={arXiv preprint arXiv:2607.27191},
  year={2026}
}

@article{priem2022openalex,
  title={{OpenAlex}: A fully-open index of scholarly works, authors, venues, institutions, and concepts},
  author={Priem, Jason and Piwowar, Heather and Orr, Richard},
  journal={arXiv preprint arXiv:2205.01833},
  year={2022}
}

@misc{field2026views,
  title        = {{AI} Researchers' Views on Automating {AI} {R\&D} and Intelligence Explosions},
  author       = {Field, Severin and Douglas, Raymond and Krueger, David},
  year         = {2026},
  eprint       = {2603.03338},
  archivePrefix= {arXiv},
  primaryClass = {cs.CY},
  doi          = {10.48550/arXiv.2603.03338},
  url          = {https://arxiv.org/abs/2603.03338}
}

@misc{chen2026rsi,
  title        = {Recursive Self-Improvement in {AI}: From Bounded Self-Refinement to Autonomous Research Loops},
  author       = {Chen, Mingguang and Wang, Licheng and Qu, Bo},
  year         = {2026},
  eprint       = {2607.07663},
  archivePrefix= {arXiv},
  primaryClass = {cs.AI},
  doi          = {10.48550/arXiv.2607.07663},
  url          = {https://arxiv.org/abs/2607.07663}
}

@misc{anthropic2026buildsitself,
  title        = {When {AI} Builds Itself},
  author       = {{The Anthropic Institute}},
  year         = {2026},
  month        = jun,
  howpublished = {Anthropic},
  url          = {https://www.anthropic.com/institute/recursive-self-improvement},
  note         = {Accessed 2026-08-18}
}

@misc{anthropic2026opus47card,
  title        = {Claude {Opus} 4.7 System Card},
  author       = {{Anthropic}},
  year         = {2026},
  month        = apr,
  howpublished = {Anthropic},
  url          = {https://www.anthropic.com/claude-opus-4-7-system-card},
  note         = {Published 16 April 2026. PDF: \url{https://www-cdn.anthropic.com/037f06850df7fbe871e206dad004c3db5fd50340/Claude\%20Opus\%204.7\%20System\%20Card.pdf}. Accessed 2026-08-18}
}

@inproceedings{chen2025mlrbench,
  title        = {{MLR-Bench}: Evaluating {AI} Agents on Open-Ended Machine Learning Research},
  author       = {Chen, Hui and Xiong, Miao and Lu, Yujie and Han, Wei and Deng, Ailin and He, Yufei and Wu, Jiaying and Li, Yibo and Liu, Yue and Hooi, Bryan},
  booktitle    = {Advances in Neural Information Processing Systems 38 (NeurIPS 2025), Datasets and Benchmarks Track},
  year         = {2025},
  eprint       = {2505.19955},
  archivePrefix= {arXiv},
  primaryClass = {cs.LG},
  doi          = {10.48550/arXiv.2505.19955},
  url          = {https://arxiv.org/abs/2505.19955}
}

@misc{ansari2026compounddeception,
  title        = {Compound Deception in Elite Peer Review: A Failure Mode Taxonomy of 100 Fabricated Citations at {NeurIPS} 2025},
  author       = {Ansari, Mohammad Samar},
  year         = {2026},
  eprint       = {2602.05930},
  archivePrefix= {arXiv},
  primaryClass = {cs.DL},
  doi          = {10.48550/arXiv.2602.05930},
  url          = {https://arxiv.org/abs/2602.05930}
}

@misc{bienz2026mysteriouscitations,
  title        = {The Case of the Mysterious Citations},
  author       = {Bienz, Amanda and Pearson, Carl and de Gonzalo, Simon Garcia},
  year         = {2026},
  eprint       = {2602.05867},
  archivePrefix= {arXiv},
  primaryClass = {cs.DL},
  doi          = {10.48550/arXiv.2602.05867},
  url          = {https://arxiv.org/abs/2602.05867}
}

@misc{openclaw,
  title        = {{OpenClaw}: Your Assistant, on Your Devices, in Your Chats},
  author       = {{OpenClaw Contributors}},
  howpublished = {GitHub repository},
  url          = {https://github.com/openclaw/openclaw},
  year         = {2026},
  note         = {MIT licence. Accessed 2026-08-18}
}

@misc{crossref_docs,
  title        = {Crossref Documentation},
  author       = {{Crossref}},
  howpublished = {Crossref},
  url          = {https://www.crossref.org/documentation/},
  year         = {2026},
  note         = {Accessed 2026-08-18}
}

@misc{arxiv,
  title        = {{arXiv}.org e-Print Archive},
  author       = {{arXiv}},
  howpublished = {arXiv},
  url          = {https://arxiv.org/},
  year         = {2026},
  note         = {Accessed 2026-08-18}
}

@inproceedings{si2025can,
  title={Can {LLMs} generate novel research ideas? {A} large-scale human study with 100+ {NLP} researchers},
  author={Si, Chenglei and Yang, Diyi and Hashimoto, Tatsunori},
  booktitle={International Conference on Learning Representations},
  volume={2025},
  pages={94003--94092},
  year={2025}
}

@incollection{simon1989scientist,
  title={The scientist as problem solver},
  author={Simon, Herbert A},
  booktitle={Complex Information Processing: The Impact of {Herbert A. Simon}},
  editor={Klahr, David and Kotovsky, Kenneth},
  publisher={L. Erlbaum Associates},
  pages={375--398},
  year={1989}
}

@book{peirce1934collected,
  title={Collected papers of {Charles Sanders Peirce}},
  author={Peirce, Charles Sanders},
  volume={5},
  year={1934},
  publisher={Harvard University Press}
}

@book{fisher1966design,
  title={The design of experiments},
  author={Fisher, Ronald Aylmer},
  edition={8th},
  year={1966},
  publisher={Oliver and Boyd},
  address={Edinburgh}
}

@incollection{gelman2016statistical,
  title={The statistical crisis in science},
  author={Gelman, Andrew and Loken, Eric},
  booktitle={The Best Writing on Mathematics 2015},
  publisher={Princeton University Press},
  pages={305--318},
  year={2016},
  doi={10.1515/9781400873371-028}
}

@article{silberzahn2018many,
  title={Many analysts, one data set: Making transparent how variations in analytic choices affect results},
  author={Silberzahn, Raphael and Uhlmann, Eric L and Martin, Daniel P and Anselmi, Pasquale and Aust, Frederik and Awtrey, Eli and Bahn{\'\i}k, {\v{S}}t{\v{e}}p{\'a}n and Bai, Feng and Bannard, Colin and Bonnier, Evelina and others},
  journal={Advances in Methods and Practices in Psychological Science},
  volume={1},
  number={3},
  pages={337--356},
  year={2018},
  publisher={Sage Publications Sage CA: Los Angeles, CA}
}

@article{steegen2016increasing,
  title={Increasing transparency through a multiverse analysis},
  author={Steegen, Sara and Tuerlinckx, Francis and Gelman, Andrew and Vanpaemel, Wolf},
  journal={Perspectives on Psychological Science},
  volume={11},
  number={5},
  pages={702--712},
  year={2016},
  publisher={Sage Publications Sage CA: Los Angeles, CA}
}

@inproceedings{chan2025mle,
  title={{MLE-Bench}: {Evaluating} machine learning agents on machine learning engineering},
  author={Chan, Jun Shern and Chowdhury, Neil and Jaffe, Oliver and Aung, James and Sherburn, Dane and Mays, Evan and Starace, Giulio and Liu, Kevin and Maksin, Leon and Patwardhan, Tejal and others},
  booktitle={International Conference on Learning Representations},
  volume={2025},
  pages={50466--50494},
  year={2025}
}

@inproceedings{tian2024debugbench,
  title={{DebugBench}: {Evaluating} debugging capability of large language models},
  author={Tian, Runchu and Ye, Yining and Qin, Yujia and Cong, Xin and Lin, Yankai and Pan, Yinxu and Wu, Yesai and Haotian, Hui and Weichuan, Liu and Liu, Zhiyuan and others},
  booktitle={Findings of the Association for Computational Linguistics: ACL 2024},
  pages={4173--4198},
  year={2024}
}

@article{chinn1993role,
  title={The role of anomalous data in knowledge acquisition: A theoretical framework and implications for science instruction},
  author={Chinn, Clark A and Brewer, William F},
  journal={Review of Educational Research},
  volume={63},
  number={1},
  pages={1--49},
  year={1993},
  publisher={Sage Publications Sage CA: Thousand Oaks, CA}
}

@article{kulkarni1988processes,
  title={The processes of scientific discovery: The strategy of experimentation},
  author={Kulkarni, Deepak and Simon, Herbert A},
  journal={Cognitive Science},
  volume={12},
  number={2},
  pages={139--175},
  year={1988},
  publisher={Elsevier}
}

@article{skarlinski2024language,
  title={Language agents achieve superhuman synthesis of scientific knowledge},
  author={Skarlinski, Michael D and Cox, Sam and Laurent, Jon M and Braza, James D and Hinks, Michaela and Hammerling, Michael J and Ponnapati, Manvitha and Rodriques, Samuel G and White, Andrew D},
  journal={arXiv preprint arXiv:2409.13740},
  year={2024}
}

@incollection{dunbar1994scientists,
  title={How scientists really reason: Scientific reasoning in real-world laboratories},
  author={Dunbar, Kevin Niall},
  booktitle={The Nature of Insight},
  publisher={The MIT Press},
  pages={365--396},
  year={1994},
  doi={10.7551/mitpress/4879.003.0017}
}

@article{platt1964strong,
  title={Strong Inference: Certain systematic methods of scientific thinking may produce much more rapid progress than others.},
  author={Platt, John R},
  journal={Science},
  volume={146},
  number={3642},
  pages={347--353},
  year={1964},
  publisher={American Association for the Advancement of Science}
}

@article{chamberlin1890method,
  title={The method of multiple working hypotheses},
  author={Chamberlin, Thomas C},
  journal={Science},
  volume={15},
  number={366},
  pages={92--96},
  year={1890},
  publisher={American Association for the Advancement of Science}
}

@book{mayo1996error,
  title={Error and the growth of experimental knowledge},
  author={Mayo, Deborah G},
  year={1996},
  publisher={University of Chicago Press}
}

@techreport{ukaisi2026securityincident,
  author      = {{UK AI Security Institute}},
  title       = {{Security Incident INC-2026-07-28-01}},
  institution = {UK AI Security Institute},
  type        = {Technical Report},
  number      = {INC-2026-07-28-01},
  year        = {2026},
  month       = jul,
  url         = {https://cdn.prod.website-files.com/663bd486c5e4c81588db7a1d/6a724858f7db25c81487016d_Security%20Incident%20INC-2026-07-28-01.pdf}
}

@techreport{oai-navier,
    author = {{OpenAI}},
    title = {Finite time blowup for {Navier--Stokes}},
    institution = {{OpenAI}},
    year = {2026},
    url = {https://cdn.openai.com/pdf/32d9f210-8b73-45e0-91bc-82a30aef8a9a/navier-stokes.pdf}
}

@incollection{laudan1983demise,
  title={The demise of the demarcation problem},
  author={Laudan, Larry},
  booktitle={Physics, philosophy and psychoanalysis: Essays in honour of Adolf Gr{\"u}nbaum},
  pages={111--127},
  year={1983},
  publisher={Springer}
}

@book{feyerabend1975against,
  title={Against method: Outline of an anarchistic theory of knowledge},
  author={Feyerabend, Paul},
  year={1975},
  publisher={New Left Books}
}

@book{polanyi1958personal,
  title={Personal knowledge: Towards a post-critical philosophy},
  author={Polanyi, Michael},
  year={1958},
  publisher={University of Chicago Press}
}

@inproceedings{bavaresco2025llms,
  title={{LLMs} instead of human judges? {A} large scale empirical study across 20 {NLP} evaluation tasks},
  author={Bavaresco, Anna and Bernardi, Raffaella and Bertolazzi, Leonardo and Elliott, Desmond and Fern{\'a}ndez, Raquel and Gatt, Albert and Ghaleb, Esam and Giulianelli, Mario and Hanna, Michael and Koller, Alexander and others},
  booktitle={Proceedings of the 63rd Annual Meeting of the Association for Computational Linguistics (Volume 2: Short Papers)},
  pages={238--255},
  year={2025}
}

@article{romeo2026exploring,
  title={Exploring automation bias in human--{AI} collaboration: a review and implications for explainable {AI}},
  author={Romeo, Giuseppe and Conti, Daniela},
  journal={AI \& Society},
  volume={41},
  number={1},
  pages={259--278},
  year={2026},
  publisher={Springer}
}

@misc{metr2026incidentinvestigation,
  author       = {{METR}},
  title        = {Brief independent investigation of agents' behavior, reasoning and collaboration in the {OpenAI} / {Hugging Face} hacking incident},
  year         = {2026},
  month        = aug,
  howpublished = {\url{https://metr.org/blog/2026-08-26-openai-hugging-face-incident-investigation/}},
  note         = {26 August. Contributors: Ryan Greenblatt, Ajeya Cotra, and Hjalmar Wijk.}
}

@misc{transluce_docent_plans,
  title        = {Analysis Plans},
  author       = {{Transluce}},
  year         = {2026},
  howpublished = {\url{https://docs.transluce.org/analysis/analysis-plans}},
  note         = {Accessed 2026-09-23}
}

@misc{meridianlabs,
  author  = {{Meridian Labs}},
  title   = {{Meridian Labs}},
  year    = {2026},
  url     = {https://meridianlabs.ai/},
  urldate = {2026-09-01}
}
